\documentclass[10pt,twocolumn,letterpaper]{article}

\usepackage{cvpr}
\usepackage{times}
\usepackage{epsfig}
\usepackage{graphicx}
\usepackage{subfigure}
\usepackage{epstopdf}
\usepackage{multirow}
\usepackage{amsmath,amssymb,amsfonts}
\usepackage{amsopn}
\usepackage{array}
\usepackage{algorithm}
\usepackage{algorithmic}
\usepackage{arydshln}
\usepackage{amsthm}

\usepackage[pagebackref=true,breaklinks=true,letterpaper=true,colorlinks,bookmarks=false]{hyperref}
\usepackage{expl3}

\cvprfinalcopy 

\def\cvprPaperID{1802} 

\ifcvprfinal\fi

\begin{document}

\title{Learning Hypergraph-regularized Attribute Predictors}


\author{Sheng Huang$^\dag$ \quad Mohamed Elhoseiny$^\ddag$ \quad Ahmed Elgammal$^\ddag$ \quad Dan Yang$^{\dag,*}$ \\
$^\dag$Chongqing University, P.R.China\quad$^\ddag$Rutgers University, USA\\
{\tt\small \{huangsheng,dyang\}@cqu.edu.cn\quad {\{mhe19,elgammal\}@cs.rutgers.edu}}}

\maketitle

\begin{abstract}
We present a novel attribute learning framework named Hypergraph-based Attribute Predictor (HAP). In HAP, a hypergraph is leveraged to depict the attribute relations in the data. Then the attribute prediction problem is casted as a regularized hypergraph cut problem in which HAP jointly learns a collection of attribute projections from the feature space to a hypergraph embedding space aligned with the attribute space.
The learned projections directly act as attribute classifiers (linear and kernelized). This formulation leads to a very efficient approach. By considering our model as a multi-graph cut task, our framework can flexibly incorporate other available information, in particular class label. We apply our approach to attribute prediction, Zero-shot and $N$-shot learning tasks. The results on AWA, USAA and CUB databases demonstrate the value of our methods in comparison with the state-of-the-art approaches.
\end{abstract}


\section{Introduction}
\vspace{-0.1cm}
Attribute learning aims at achieving an intermediate representation on top of the low-level visual feature space, which encodes semantic properties shared across different categories of objects or scenes. Such an intermediate representation plays a role as the vehicles of semantics in human-machine communication.
Farhadi \etal\cite{objatt} and Lampert~\etal\cite{dap} showed that supervised attributes can be transferred across object categories, allowing description and naming of objects from categories not seen during training. Therefore, attributes provide a way to encode and share knowledge to achieve challenging tasks such as the Zero-Shot Learning (ZST) problem, where the goal is to categorize classes that are unseen during training~\cite{dap,dap2014}. 

A lot of approaches have been proposed for attribute learning, \eg\cite{dap,objatt,attret,parkash2012attributes,labelemb,decor,lmla,ra}. Two fundamental issues remain unsolved, although recent researches have started to pay attention to them \eg\cite{labelemb,decor,lmla}. First, traditional approaches learn attributes independently from each other (one-vs-all classifiers)~\cite{dap,objatt}, without explicitly exploiting the correlation between attributes. Second, learning attribute classifiers are typically done independent of the subsequent tasks, such as categorization or zero shot learning, and typically category labels are ignored in the learning process. Optimizing the attribute prediction independent of the succeeding task does not guarantee to yield the best attribute predictor for that task.

\begin{figure}[!tbp]
\setlength{\abovecaptionskip}{-0.1cm}
\setlength{\belowcaptionskip}{-0.1cm}
\centering{\includegraphics[scale=0.4]{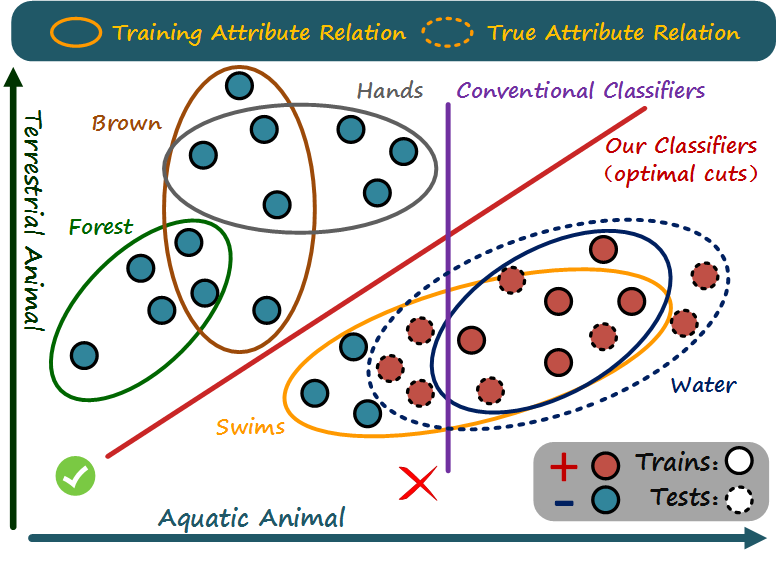}}
\caption{The visualization of the hypergraph cut for predicting the attribute "water". Each ellipse with solid lines denotes a hyperedge that encodes an attribute relation and each circle represents a sample. }
\label{hypercut}
\vspace{-0.5cm}
\end{figure}

Correlations naturally exist among attributes. Is it better to exploit the correlation or to discourage the correlation during attribute learning, \ie decorrelation? Several papers have argued that exploiting correlation between attributes improves their discriminative powers, \eg\cite{dla,attface}.
Recent works attempted to address the issue of joint attribute learning  with a focus on decorrelating attributes~\cite{labelemb,decor}. Jayaraman \etal\cite{decor} argued that attribute learning approaches are prone to learn visual features that correlates with attributes, not attributes themselves. Therefore they argued for decorrelation of attributes, by exploiting feature competition during learning through a multitask learning framework. Decorrelation of attributes might be suitable in tasks such as describing images with text, key-word based retrieval, or generating image annotations. However, preserving and exploiting correlation between attributes should preserve the natural clustering in the data and should be better for classification and zero-shot learning tasks. Correlation is a nature of attributes and compulsively decorrelating the attributes may break the original relations of attributes in the visual space as well as in the semantic space.

 To illustrate our point we use the example in Figure~\ref{hypercut} where we used attributes from the AWA dataset~\cite{dap}. Consider learning the attribute ``water''. In the figure, there are two clusters denoted as two superclasses (terrestrial and aquatic animals). It is expected that the attributes in each cluster will be highly correlated and coexist in images. The conventional classifiers directly learn an optimal separation for the attribute ``water'' which clearly ignores the correlations as well as the natural clusters in the data. Although it achieves the optimal attribute prediction, it clearly reduces the utility of attribute predictors in  subsequent tasks such as categorization, and is much easier to get mired in overfitting. Instead we aim at a cut that, besides minimizing the attribute prediction loss, tries to preserve the clustering in the data. Preserving the correlation can be even more beneficial for attributes that are not visual. For example the attribute ``swim'' describes an action and it is very hard to predict it from visual features if it is forced to be decorrelated from other attributes such as ``water''.

\begin{figure}[!tbp]
\setlength{\abovecaptionskip}{-0.1cm}
\setlength{\belowcaptionskip}{-0.1cm}
\centering{\includegraphics[scale=0.18]{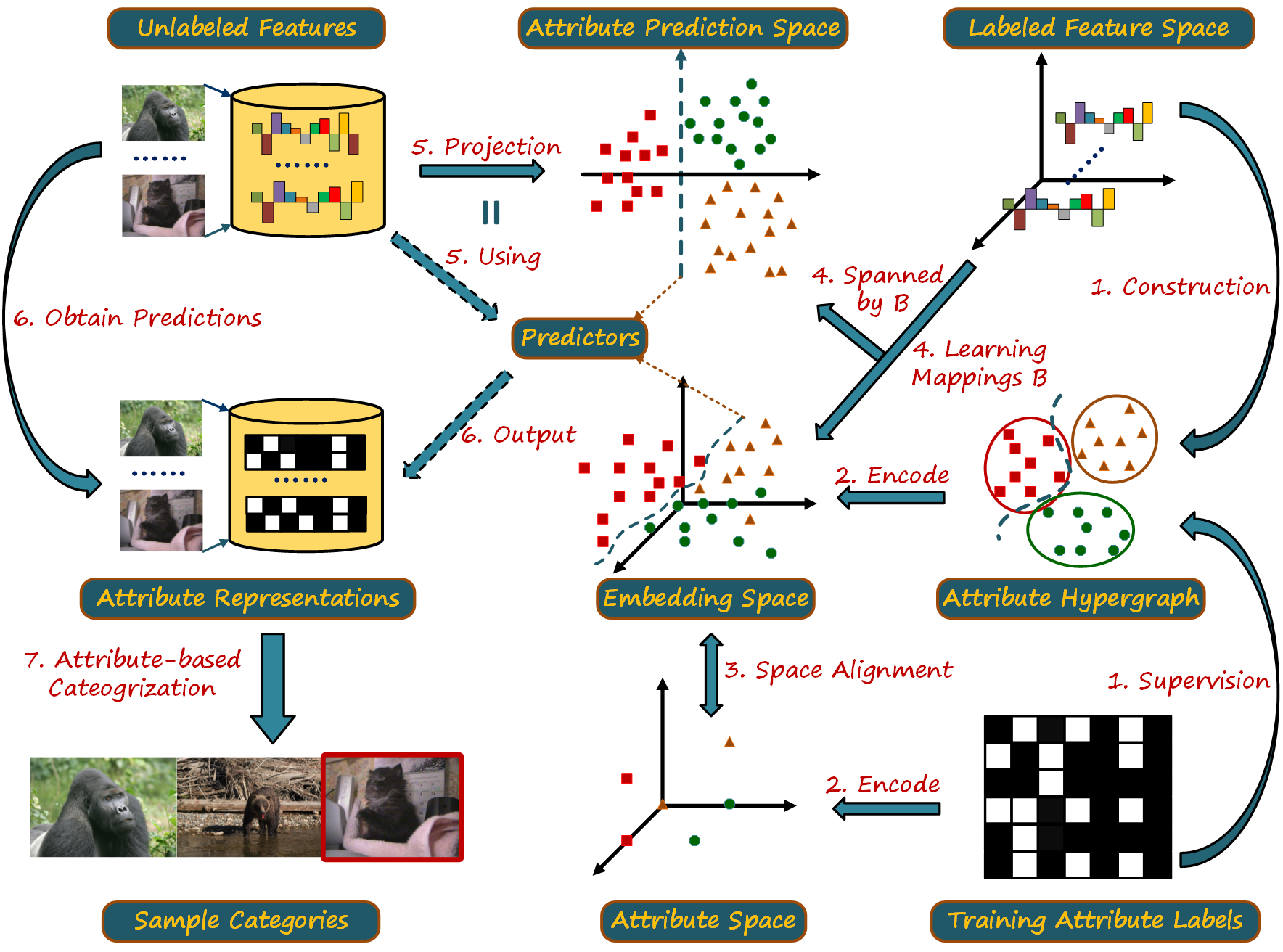}}
\caption{The overview of our approach, we learn a collection of the mappings (projections) from the feature space to the hypergraph embedding space which is aligned by the attribute space and encode the attribute relations. The learned projections span the Attribute Prediction Space (APS) in which each basis is an attribute predictor. The attribute prediction of a sample can be achieved by projection it to the APS and then the attribute-based categorization can be performed. }\label{aps}
\vspace{-0.5cm}
\end{figure}

Our goal is to design a new attribute learning framework that addresses the two aforementioned issues, \ie jointly learning attributes while exploiting the correlation between attributes, and exploiting class information as well as any available side information. We propose to model the attribute learning as a supervised hypergraph cut problem. As a generalization of graphs, hypergraphs are typically used to depict the high-order and multiple relations of data~\cite{hyper,hyperclass,hyperml,hsl}. One merit of hypergraph is that it can capture the correlations of multiple relations, since the partition of a vertex set who has many common hyperedges will lead to a heavy penalty~\cite{hsl}. In our formulation, we define a hypergraph where each vertex  corresponds to a sample and a hyperedge is a vertex set sharing the same attribute label. Then, we can consider the attribute prediction problem as a hypergraph cut problem. More specifically, a collection of hypergraph cuts (one cut per attribute) that minimizes the loss of attribute relations (defined by the hyperedges) is jointly learned.  Moreover, such cuts also minimize the attribute prediction errors of training data.  Since hypergraph cuts can be deemed as the hypergraph embedding from the perspective of graph embedding~\cite{ncut,lapeig,graphemb}, this step actually tries to align the embedding space, which encodes the attribute relations, with the semantic attribute space. We also propose attribute predictors (or classifiers) that can be obtained by introducing a mapping from the feature space to this aligned hypergraph embedding space. We name this approach Hypergraph-based Attribute Predictor (HAP), which can be combined with different hypergraph models to obtain classifiers from the cuts. We illustrate our model in Figure~\ref{aps}.

In order to incorporate class information and any additional side information within the HAP formulation, we consider it as a multi-graph cut problem. One or several additional graphs (or hypergraphs), which encode side information, will be introduced as the penalties to the HAP. In that case, the new cuts should not only minimize the loss of attribute relations, but also the losses of the side information. In case of class labels, we formulate a new HAP framework, denoted as Class-Specific HAP (CSHAP), which enhances the discriminating ability of the attribute predictors. A hypergraph and a graph, which encode the class information in two different ways, are respectively leveraged to produce two different versions of the CSHAP approaches. We denote them Hypergraph-based CSHAP (CSHAP$_H$) and Graph-based CSHAP (CSHAP$_G$) respectively. Finally, all the proposed approaches are kernelized to incorporate the nonlinearity.

We summarize the  contributions of this paper as follows:
\vspace{-0.6cm}
\begin{enumerate}
 \setlength{\itemsep}{0pt}
 \setlength{\parskip}{0pt}
 \setlength{\parsep}{0pt}
 \item As far as we know, our approach is the first to formulate attribute learning as a supervised hypergraph cut problem.
 \item We propose an approach to construct predictors (linear and nonlinear classifier) jointly while solving for the cuts. This idea is applicable in general (not limited to the context of attribute learning) to any hypergraph cut algorithm, as a general path to derive the classifiers from graph model.
 \item We provide a flexible solution to incorporate the side information in attribute learning.
 \item The proposed approach provides efficient attribute predictions, since the computational complexity of the attribute prediction is linear with respect to the dimension of feature.
\end{enumerate}
\vspace{-0.2cm}

We experimented with three datasets: Animal With Attributes (AWA)~\cite{dap}, Caltech-UCSD Birds (CUB)~\cite{cub} and Unstructured Social Activity Attribute (USAA)~\cite{lmla}. The results on attribute prediction, Zero-shot, N-shot Learning, and categorization consistently validate the effectiveness of the proposed framework.
The rest of paper is organized as follows: Section \ref{s2} presents the related works; Section \ref{s3} describes the proposed approach. Section \ref{s4} shows the experimental evaluation of our works; the conclusion is summarized in Section \ref{s6}.

\vspace{-0.15cm}
\section{Related Works}\label{s2}
\vspace{-0.1cm}
\subsection{Attributes}
\vspace{-0.1cm}

Traditional attribute learning approaches follow a supervised discriminative learning pipeline (one-vs-all classifiers) where attribute classifiers are learned independently given attribute labels for each image or each class~\cite{objatt,dap}. Recently several papers suggested approaches for joint learning of attributes~\cite{joint,dla,labelemb,decor}. Wang \etal~\cite{dla} and Song~\etal~\cite{attface} construct a graph of attributes in attribute domain from the training data and consider it as the latent variables in the latent SVM for categorization, and they showed that exploiting attribute relations helps improve class prediction. In contrast our hypergraph is constructed on the samples, which facilitates aligning the feature space with the attribute semantic space.

Mahajan \etal\cite{joint} proposed a joint learning framework that removes the correlations as the redundancies during learning the mapping between attributes and classes which actually ignores the contributions of the co-occurrence attributes.  Akata \etal\cite{labelemb} proposed an approach that simultaneously target three problems: optimizing attribute prediction using class labels, using side information, and incremental learning. They achieved decorrelation by using dimensionality reduction on the class-attribute matrix, \ie in the label space, and showed that the attribute dimensions can be reduced significantly without affecting the accuracy. In contrast our hypergraph construction achieve correlation/decorrelation by employing the sample-attribute relations and embeding the data samples in a space that is aligned with the attribute labels. Recently, Jayaraman \etal~\cite{decor} attempted to decorrelate the attributes via solving a structure sparsity model in which semantic attribute groups are manually provided as auxiliary data.


Attribute learning is just the preliminary step for some other visual tasks. Few approaches have been proposed to incorporate the additional information, in particular class labels into attribute learning for benefiting the subsequent tasks~\cite{objatt,labelemb,dla,iaa,cla}. However, the attribute learning and the exploitation of the additional information are highly coupled in these approaches. It is very hard to add novel additional information into the model, or unplug the additional information exploitation part from the original models when the additional information of the given data is not available. In contrast, our proposed method can flexibly to address this issue by considering the attribute prediction task as multi-graph (hypergraph) cut problem, enabling adding any side information as an extra graph or hypergraph.

Zero-Shot Learning (ZSL) is the task of object recognition for categories with no training examples~\cite{dap}. Several intermediate representations has been used for ZSL, including attributes~\cite{dap,objatt,dap2014,labelemb}, linguistic knowledge~\cite{tmve,ekt}, textual description~\cite{wac} and visual abstraction~\cite{zslva}.
The core attribute-based ZSL approaches are the attribute learning. Therefore our attribute prediction can be integrated to several ZSL frameworks. Although our work is an attribute-based ZSL, other intermediate representation can also be readily plugged into the HAP framework to replace the attributes, since it generally provides a mapping from the low-level representation to intermediate representation.  Several ZSL approaches can be extended to N-Shot Learning~\cite{zol,labelemb,tmve,aar}.

\vspace{-0.05cm}
\subsection{Hypergraph Learning}
\vspace{-0.15cm}
Hypergraph is a generalization of the regular graph which has been widely applied to depict the high-order relations between data points~\cite{hyper,hyperclass,hyperml,hsl}. Since the attribute prediction task can be regarded as a multi-label classification problem, we will introduce not only the relevant hypergraph model, but also some hypergraph-based multi-label classification algorithms, which motivated our work. More specifically, Zhou \etal~\cite{hyper} proposed a normalized hypergraph model for embedding and transduction. Our formulation is based on Zhou's model, however it is inductive.  Moreover, we show how the hypergraph cut can be reformulated to provide direct linear or nonlinear class predictors.
Chen \etal~\cite{hyperclass} leveraged the hypergraph to capture the correlation of categories and introduced it as a regularization to SVM model for multi-label classification. Similar to~\cite{hyperclass}, Sun \etal~\cite{hsl} used the hypergraph to capture the correlation of classes and performed a hypergraph embedding as the new representation for multi-class classification. In~\cite{hyperml} the hypergraph is utilized to measure the loss of multi-labels during the multi-kernel learning.
In contrast to all the existing hypergraph learning algorithms, as far as we know, our model is the first approach that directly derives the multi-label classifier from the hypergraph embedding.

\vspace{-0.15cm}
\section{Approach}\label{s3}
\vspace{-0.1cm}
\subsection{Preliminaries}
\vspace{-0.1cm}
We start by reviewing some basic definitions of hypergraphs and introducing the notations. Hypergraphs are a generalization of graphs in which a hyperedge (the analogy of an edge) is an arbitrary non-empty subsets of the vertex set~(Fig~\ref{hypercut} shows a hypergraph with five hyperedges). Given a hypergraph $G=(V,E)$ in an arbitrary feature space, $V$ and $E$ are the vertex set and hyperedge set, where each vertex and hyperedge are respectively defined as $v\in V$ and $e \in E$. Moreover, a hyperedge $e$ is a subset of $V$. The vertex-edge incidence matrix $H\in \mathcal{R}^{|V|\times |E|}$ is defined as follows
\begin{equation}\label{}
\small
h(v,e)=\left\{
\begin{array}{l l}
{1,~\text{if}~v\in e}\\
{0,~\text{otherwise}}.
\end{array}
\right.
\end{equation}
The degree of a hyperedge $e$, which is denoted as $\delta(e)$, is the number of vertices in $e$
\begin{equation}\label{}
\small  \delta(e)=\sum_{v\in e}h(v,e),
\end{equation}
and the degree of a vertex $v\in V$ is defined as follows
\begin{equation}\label{}
\small  d(v)=\sum_{v\in e, e\in E} w(e)=\sum_{e\in E} w(e)h(v,e),
\end{equation}
where $w(e)$ is the weight of the hyperedge $e$. We denote the diagonal matrix forms of $\delta(e)$, $d(v)$ and $w(e)$ as $D_e$, $D_v$ and $W$ respectively. Note, here we defined $H$ as a binary matrix for simplicity. For the continuous value case, a probabilistic hypergraph model~\cite{phyper} can be adopted in which each element of $H$ denotes the probability of a vertex in a hyperedge.
\vspace{-0.1cm}
\subsection{Attribute Hypergraph}
\vspace{-0.1cm}
In our model, we define a hypergraph to depict the attribute relations of samples (corresponding to images in the training set). In this hypergraph, the vertex $v_i\in V$ is corresponding to the sample $x_i \in X$, which is the $i$-th column of the $d\times n$-dimensional sample matrix $X$. Here, $n$ is the number of samples and $d$ is the dimension of the feature space. Each hyperedge is defined as a vertex set that shares the same attribute label. In such case, the number of hyperedges is equal to the number of attributes $m$ and the $n\times m$-dimensional matrix incident matrix $H$ is exactly the attribute label matrix.
The more common attributes among a set of images, the more hyperedges will exist between their corresponding vertices, and the stronger the link will be between these vertices.
Therefore, break such a link will lead to a heavy penalty during the learning process. In this way, the hypergraph actually provides a natural way to capture the correlation/decorrelation of attributes. We regard the hyperedge $e$ as a clique and consider the mean of the heat kernel weights of the pairwise edges in this clique as the hyperedge weight
\begin{equation}\label{}
  \small w(e)= \frac{1}{\delta(e)(\delta(e)-1)}\sum_{\{v_i,v_j\}\in e} \exp\left(-\frac{||x_i-x_j||^2}{\mu}\right).
\end{equation}
Certainly, some other hyperedge weighing schemes can be also applied.
\vspace{-0.1cm}
\subsection{Hypergraph-based Attribute Predictor}
\vspace{-0.1cm}
Normalized hypergraph cut is often utilized to learn the high-order relation and correlation information. The main idea of our model stems from the hypergraph-based transduction which is regarded as a regularized normalized hypergraph cut model~\cite{hyper}. In contrast, our method is a supervised inductive model. Since our proposed attribute predictor (or classifier) is based on the hypergraph model, we call it Hypergraph-based Attribute Predictor (HAP).


In HAP, a collection of hypergraph cuts $F=[f_1,\cdots,f_m]$ is defined as the predictors of attributes in the feature space where $m$ is the number of attributes and the cut $f_i$ is a column vector whose elements are the predictions of $i$-th attribute for each sample. An optimal cut should not disrupt the hyperedges during hypergraph partition as much as possible. In other words, the optimal cut should keep the attribute relations of samples as much as possible since each hyperedge is given by an attribute label. Similar to Zhou's normalized hypergraph~\cite{hyper}, we can define an attribute relation loss function with respect to the given hypergraph $G$ and a collection of hyperedge cuts $F$ can be denoted as follows
\begin{equation}\label{oinfloss}
 \small
    \Omega(F, G)=\frac{1}{2}\sum_{e\in E}\sum_{(u,v)\in e}\frac{w(e)}{\delta(e)}\left|\left|\frac{F_u}{\sqrt{d(u)}}-\frac{F_v}{\sqrt{d(v)}}\right|\right|^2,
\end{equation}
where $F_u$ returns a row vector corresponding to the predictions of attributes for the vertex $u$. Clearly, the loss will be reduced when the signs of $F_u$ and $F_v$ are identical. Following some deductions, Equation~\ref{infloss} can be reformulated as follows
{\small
\begin{eqnarray}\label{infloss}\nonumber
  \Omega(F, G)&=&\frac{1}{2}\sum_{e\in E}\sum_{(u,v)\in e}\frac{w(e)}{\delta(e)}\left|\left|\frac{F_u}{\sqrt{d(u)}}-\frac{F_v}{\sqrt{d(v)}}\right|\right|^2\\\nonumber
  &=&\text{Tr}(F^T(I-D_v^{-1/2}{HW D_e^{-1}H^T}{D_v^{-1/2}})F)\\
  &=&\text{Tr}(F^TL_HF),
\end{eqnarray}
}where $L_H$ is the normalized hypergraph Laplacian matrix which is derived from the hypergraph of attributes, and $I$ is an identity matrix. $\text{Tr}(\cdot)$ is the trace of the matrix. The detail deductions of Equation~\ref{infloss} can be found in the supplementary material.

Besides measuring the loss of attribute relation information, we also need to consider the attribute prediction errors of the train data, which can be obtained via calculating the Euclidean distance between attribute predictions $F$ and the  attribute label matrix. In order to make zero as the classification boundary, we define a shifted attribute label matrix $Y$ via shifting the attribute labels as $Y=2H-\mathbf{1}$ where $\mathbf{1}$ is a matrix of the same size as $H$ whose elements are all equal to 1. In $Y$, if an attribute exists in a sample, its corresponding attribute label is 1, otherwise it is -1. Given this definition, the attribute prediction loss is defined as
\begin{equation}\label{predloss}
  \small
  \Delta(F,Y)=||F-Y||^2
\end{equation}
Simultaneously minimizing the the previous two losses leads to our model
{\small
\begin{eqnarray}\label{predmodel}\nonumber
  \hat{F}&=&\arg\underset{F}\min\{\Omega(F, G)+\lambda \Delta(F, Y)\}\\
  &=&\text{Tr}(F^TL_HF)+\lambda ||F-Y||^2,
\end{eqnarray}
}where $\lambda$ is a positive parameter to reconcile these two losses. Now, the optimal hypergraph cut $f_i$ introduces a binary partition to hypergraph that can preserve the information of $i$-th attribute relation and reduce the prediction error of $i$-th attribute as much as possible.

From the perspective of graph embedding, the hypergraph cuts are the embedding of the given hypergraph, where the embedding coordinate of sample $u$ is the $u$-th row of $F$. Equation~\ref{predmodel} actually aligns the hypergraph embedding space (defined by $F$) with the shifted attribute space. Consequently, we now transform the problem of seeking attribute predictors/cuts to the problem of finding a mapping from the feature space to this aligned embedding space, \ie
\begin{equation}\label{proj}
\small
  F =X^TB,
\end{equation}
where the projection matrix $B=[\beta_1,\cdots,\beta_i,\cdots,\beta_m]$ is such collection of mappings whose $i$-th column is a predictor of the $i$-th attribute, corresponding to  the $i$-th hypergraph cut $f_i=\beta_i^T X$. We then substitute Equation~\ref{proj} into Equation~\ref{predmodel}, and introduce $L_2$-norm constraint to $B$ to avoid the overfitting. Thus, the Equation~\ref{predmodel} is reformulated as the following optimization problem with respect to $B$
\begin{equation}\label{modelB}
\small
\hat{B}=\arg\underset{B}\min(\text{Tr}(B^TXL_HX^TB)+\lambda ||X^TB-Y||^2+\eta ||B||^2).
\end{equation}
where $\eta$ is a positive regularization parameter. Since $L_H$ is a positive semi-definite matrix, this problem is a typical Regularized Least Square (RLS) problem that can be efficiently solved. We obtain the partial derivative of Equation~\ref{modelB} with respect to $B$, and equate it to zero, which leads to a closed-form solution for $B$ as follows
{\small
\begin{eqnarray}\label{sols} \nonumber
&&XL_hX^TB+\lambda(XX^T-XY)+\eta B=0\\
&\Rightarrow&~B=(XL_HX^T+\lambda XX^T+\eta I)^{-1}(\lambda XY)\\ \nonumber
&\Rightarrow&~B=(XL_HX^T+\lambda XX^T+\eta I)^{-1}(\lambda X(2H-\mathbf{1})).
\end{eqnarray}
}At test time, given a unlabeled sample $z_i$, its attribute predictions can be achieved by projecting the sample into the subspace spanned by $B$,
\begin{equation}\label{}
\small
  \mathbf{p}_i=\text{sign}({z_i^TB}),
\end{equation}
where $\text{sign}(\cdot)$ returns the sign of each element of a vector and $\mathbf{p}_i=[p_{i1},\cdots,p_{ij},\cdots,p_{im}]$ is a row vector encoded the predicted attributes. Its $j$-th element $p_{ij}=z_i^T\beta_j$ is the confidence of the existence of the $j$-th attribute with respect to the sample $z_i$. We call the subspace spanned by $B$ Attribute Prediction Space (APS), since each basis of this space actually is a predictor of a specific attribute.

\vspace{-0.1cm}
\subsection{Incorporating Class and Side Information}
\vspace{-0.1cm}
As a regularized graph learning approach, it is flexible to introduce other meaningful constraints to further enhance attribute learning. In this section, we take the class label as an example to show how to leverage any additional information to enhance our model. The exploitation of class labels can enhance the classification abilities of HAP algorithms, since homogenous samples always share more similarities in attributes. 

We adopt two approaches to incorporate the class information. The first approach uses  a hypergraph $G_C=(V,E_C)$  to depict the class relation of samples, similar to the way we used a hypergraph to depict the attribute relations in the previous subsection. It is not hard to derive the hypergraph Laplacian $L_C$ from this hypergraph via following the same way as Equation~\ref{infloss}.
The second approach, following \cite{lapeig,lpp}, constructs a pairwise graph $G_L=(V,E_L)$ in a supervised way for encoding the class information. We can encode class information using a graph, since unlike the attributes, the classes are disjoint. Two samples are connected with an edge if they belong to the same class (homogenous samples). Similar to the hypergraph model, the heat kernel weighting is adopted as the edge weighting scheme. Finally, the well known Laplacian Eigenmapping model can easily derive the graph Laplacian $L_L$.

Introducing such class-label graph $G_L$or hypergraph $G_C$ to the loss function in Equation~\ref{infloss} leads to the new loss function
{\small
\begin{eqnarray}\label{}\nonumber
   \Omega(F, G, G_*)&=&\text{Tr}(F^TL_HF+\gamma F^TL_*F)\\ \nonumber
   &=&\text{Tr}(B^TX(L_H+\gamma L_*)X^TB)\\
   &=&\text{Tr}(B^TXL_WX^TB),
\end{eqnarray}
}where $G_*$ denotes either $G_L$ or $G_C$ and $L_*$ is the corresponding Laplacian matrix. $L_W=L_H+\gamma L_*$ is the combination of the original and new Laplacian matrices.
According to Equation~\ref{modelB}, the new objective function can be reformulated as follows
\begin{equation}\label{modelBC}
\small
\hat{B}=\arg\underset{B}\min(\text{Tr}(B^TXL_WX^TB)+\lambda ||X^TB-Y||^2+\eta ||B||^2)
\end{equation}
and the solution of $B$  achieved by just replacing $L_H$ with $L_W$ in Equation~\ref{sols}
\begin{equation}\label{}
\small
B=(XL_WX^T+\lambda XX^T+\eta I)^{-1}(\lambda X(2H-\mathbf{1})).
\end{equation}
We call these models Class Specific Hypergraph-based Attribute Predictor (CSHAP). To distinguish the Hypergraph-based CSHAP and Graph-based (Laplacian eigenmapping-based) CSHAP, we respectively denote them by CSHAP$_H$ and CSHAP$_{G}$ for short. We hypothesize that  CSHAP$_{G}$ is expected to capture the intra-manifold structure between the samples better than CSHAP$_H$, since CSHAP$_H$ just group the homogenous samples using hyperedges, while CSHAP$_{G}$ preserves the pair-wise structure.

If more additional information are available, the graph Laplacians that encode these information, can be also added, $L_W=L_H+\gamma_1L_1+\cdots+\gamma_aL_a$. Such positive regularization parameters $\gamma_i,i\in\{1,\cdots,a\}$ can be deduced by multiple kernel learning, since each Laplacian matrix is associated with an affinity matrix (similarity matrix) which can be considered as a kernel matrix.

\vspace{-0.1cm}
\subsection{Kernelization of HAP}
\vspace{-0.1cm}
The mapping from the feature space to the shifted attribute space (aligned embedding space) may be not linear. This motivated us to present the kernelization for our method. According to the generalized representer theorem \cite{rth01},  a minimizer of a regularized empirical risk function over a RKHS can be represented as a linear combination of kernels, evaluated on the training set. Inspired by the representer theorem on the attribute classification risk function, we embed kernel representation of the samples (\ie $K_X B$). This transformation could be interpreted that each dimension is the embedding is linear combination of the kernel-evaluations on the training set, which matches the representer theorem.
\begin{equation}\label{}
 \small
  F=K_X B
\end{equation}
where $K_X$ is an $n \times n$ kernel matrix associated with a kernel function $k(\cdot,\cdot)$, and $n$ is the number of points in the training set. Therefore, the objective functions of the Kernelized HAP (KHAP) and Kernerlized CSHAP (KCSHAP) can be denoted as follows
{\small
\begin{eqnarray}\label{KmodelBC}\nonumber
\hat{B}=\arg\underset{B}\min\{\text{Tr}(B^T K_X L_A K_X B)+\lambda ||K_X B-Y||^2+\eta ||B||^2\}
\end{eqnarray}
}where $L_A$ is equal to $L_H$ in the KHAP case and $L_W$ in the KCSHAP case. The solution of Equation~\ref{KmodelBC} is
\begin{equation}\label{}
\small
B=(K_X L_A K_X) ^T+\lambda (K_X^2 +\eta I)^{-1}(\lambda K_X Y)
\end{equation}
Then, the attribute predictions can be obtained as follows
\begin{equation}\label{}
  \small
  \mathbf{p}(z_*)=\text{sign}({\mathbf{k}(z_*)^TB}) = \text{sign}\left(\sum_{i=1}^{N} {\mathbf{k}(z_*, x_i)^TB}\right)
\end{equation}
where $\mathbf{k}(z_*) = [k(z_*, x_1), \cdots, k(z_*, x_n)]$. $z_*$ is the test sample.

\vspace{-0.1cm}
\subsection{Zero-Shot and N-Shot Learning}
\vspace{-0.1cm}
Typically there are two ways to annotate the samples using attributes, either to assign the attributes for each sample, or to assign the attributes for each class. Our proposed approach supports both of these two scenarios. 

At zero-shot or N-shot time, before we classify samples based on the predicted attributes, we use the sigmoid function to normalize the obtained attribute confidences $\mathbf{s}_i=z_i^TB$ into the range $[0,1]$.
\begin{equation}\label{}
\small  \mathbf{r}_i=\frac{1}{1+\exp(-\frac{\mathbf{s}_i}{\rho})},
\end{equation}
where $\rho$ is a positive scaling parameter and $\mathbf{r}_i=[r_{i1},\cdots,r_{im}]$ is the normalized attribute confidence vector which can be deemed as the probabilities of the existences of attributes.

In the case where only the classes are labeled with attributes, we follow the approach of Direct Attribute Prediction (DAP)~\cite{dap,dap2014} where the Bayes' rule is adopted to calculate the posterior of a test class of a given sample based on its attribute probabilities $\mathbf{r}$. The sample is labeled with the class with the maximum posterior.

With regard to the case where each sample is annotated with attributes, we define the mean of the attribute prototypes in the same class as the attribute template for this class. We denote the template of class $j$ as $\mathbf{t}_j$. The elements of this template indicate the prior probabilities of the attributes with respect to this class. We classify the samples by directly measuring the Euclidean distance between the attribute existence probabilities of the sample and the attribute template of a class

\begin{equation}\label{ptempd}
\small
\mathcal{L}(z_i)=\arg\underset{j}\min||\mathbf{r}_i-\mathbf{t}_j||^2
\end{equation}
where $\mathcal{L}(\cdot)$ returns the class label of a sample.

\vspace{-0.1cm}
\section{Experiments}\label{s4}
\vspace{-0.1cm}
\subsection{Experimental Setups}\label{s4_1}
\vspace{-0.1cm}
\noindent{\bf Datasets:} We use three datasets to validate the proposed approach: Animal With Attributes (AWA)~\cite{dap}, Caltech-UCSD Birds (CUB)~\cite{cub} and Unstructured Social Activity Attribute (USAA)~\cite{lmla}. AWA contains 30,475 images of 50 animal classes. Each class is annotated with 85 attributes. Following~\cite{dap,dap2014}, we divide the dataset into 40 classes (24,295 images) to be used for training and 10 classes
(6180 images) for testing. CUB (2011 version)~\cite{cub} contains roughly 11,800 images of 200 bird classes. Each class is annotated with 312 binary attributes. We split the dataset following~\cite{labelemb} to facilitate direct comparison (150 classes for training and the rest 50 classes for testing). USAA is a video dataset~\cite{lmla} with 69 instance-level attributes for 8 classes of complex social group activity videos from YouTube. Each class has around 100 training and testing videos respectively. We follow~\cite{lmla} for  splitting the dataset by randomly dividing the 8 classes into two disjoint sets of four classes each  for training and testing (the mean accuracies will be reported).

\noindent{\bf Features:} We adopt the 4096-dimensional deep learning features named DeCAF~\cite{decaf} as the baseline feature for the AWA dataset since these features are already been available online for comparison\footnotemark[1]. We extract 4096-dimensional deep learning features called Caffe~\cite{caffe} for representing the images in CUB database. The USAA databases already provided the 14,000-dimensional baseline features\footnotemark[2] which are constructed from six histogram features, namely RGB color histograms, SIFT, rgSIFT, PHOG, SURF and local self-similarity histograms~\cite{lmla}.
\footnotetext[1]{http://www.ist.ac.at/~chl/AwA/AwA-features-decaf.tar.bz2.}
\footnotetext[2]{http://www.eecs.qmul.ac.uk/~yf300/USAA/download/}

\noindent{\bf Metrics:} We report the classification accuracy (in \%) averaged over the classes as the N-shot learning and ZSL accuracy in the AWA and CUB databases. In the USAA database, we follow~\cite{lmla,tmve} and report the absolute classification accuracy of data. For attribute prediction accuracies, we report the average Area Under Curve (AUC) for the ROC.

\begin{table}[!tbp]
\caption{Average Attribute Prediction Accuracies (in AUC).  \label{attpred}}
\vspace{-0.2cm}
\footnotesize
\begin{center}
    \begin{tabular}{|c|ccc|}
    \hline
     \multirow{2}*{Approaches}
    &\multicolumn{3}{c|}{Prediction Accuracies (\%)}\\ \cline{2-4}
&AWA&USAA&CUB\\
\hline
\textbf{HAP}&74.0&61.7$\pm$1.3&68.5\\
\textbf{CSHAP$_H$}&74.0&\textbf{62.2$\pm$0.8}&\textbf{68.7}\\
\textbf{CSHAP$_G$}&\textbf{74.3}&61.8$\pm$1.8&68.5\\
DAP~\cite{dap2014}&72.8/63.0$^*$&---&61.8\\
IAP~\cite{dap2014}&72.1/73.8$^*$&---&---\\
ALE~\cite{labelemb}&65.7&---&60.3\\
      \hline
    \end{tabular}{}
    \end{center}
    \vspace{-0.8cm}
\end{table}

\footnotetext[3]{We use the code and features available in the AWA webpage. The parameters of the model are well tuned using cross validation to get the best performance.}

\vspace{-0.1cm}
\subsection{Attribute Prediction}
\vspace{-0.1cm}
We report the attribute prediction performance of different approaches in Table~\ref{attpred}. Three well known attribute learning approaches, namely Direct Attribute Prediction (DAP)~\cite{dap,dap2014}, Indirect Attribute Prediction (IAP)~\cite{dap,dap2014} and Attribute Label Embedding (ALE)~\cite{labelemb} are reported for comparison. The sign '$*$' indicates the performance of running the code provided by the authors on the DeCafe features we are using, which are also provided by the authors\footnotemark[3].
From the results, we can find that HAP, CSHAP$_H$ and CSHAP$_G$ outperform all the compared approaches. For example, the accuracy gains of HAP, CSHAP$_H$ and CSHAP$_G$ over DAP are 11\%, 11\% and 11.3\% respectively under the same features and experimental settings. The CSHAP$_H$ performed slightly better than the other two HAP algorithms. This is not surprising, since the contribution of the class label in  CSHAP$_H$ and CSHAP$_G$ is expected to be limited for attribute prediction.

\vspace{-0.1cm}
\subsection{Zero-Shot Learning}\label{zsl}
\vspace{-0.1cm}
The results of seven recent ZSL approaches are complemented for comparison in Table~\ref{zslres}. These are Attribute Hierarchical Label Embedding (AHLE)~\cite{labelemb}, Hierarchies Label Embedding (HLE)~\cite{labelemb}, Multi-modal Latent Attribute Topic Model (M2LATM)~\cite{lmla}, Propagated Semantic Transfer (PST)~\cite{tlts}, Zero-Shot Random Forests (ZSRF)~\cite{zsrf}, Category-Level Attribute approach (CLA)~\cite{cla} and Decorrelated Attributes (DA)~\cite{decor}. For~\cite{zsrf,cla}  we only compared with their results on the attributes provided by the dataset, and not their results using discovered attributes. As before, the sign '$*$' indicates the performance of running the code provided by the AWA authors on the same features we are using, which are also provided by the authors.

From the results in Table~\ref{zslres}, we can notice that the proposed approaches outperform the compared methods on the AWA and USAA datasets,. For example, the accuracy gains of CSHAP$_H$ over DAP and AHLE are 2.8\% and 2.1\% on AWA dataset. On the USAA dataset, the performance improvements of HAP algorithms over other approaches are more significant. CSHAP$_H$ obtains 10.1\% and 3.0\% more accuracies in comparison with DAP and M2LATM. Although HAP algorithms have not obtained the best performance in comparison with AHLE in CUB dataset, it still outperforms other approaches. Moreover, the performance gaps between HAP algorithms and AHLE are only 0.5\% in this dataset.

In the experiments, CSHAP$_H$ often achieves better results than HAP, since the CSHAP$_H$ attempts to leverage the class labels for clustering the homogenous samples together in the attribution prediction step. Similar phenomenon can be also observed for CSHAP$_G$ while its improvement is less significant. We attribute it to the mechanism of CSHAP$_G$ where it tries to preserve the manifold structure of each class using the given class labels. In ZSL case, the test classes are unseen in the train dataset. Therefore, CSHAP$_G$ may not capture the manifold structures of unseen classes. Another interesting phenomenon is that CSHAP can enhance HAP in the AWA and USAA databases, while cannot improve on the CUB database. This is because the CUB databset has more attributes which are already enough for distinguish classes while the other two datasets have less attributes so that they benefitted more from the complementary information (The number of attributes of USAA, AWA and CUB dataset are 69, 85 and 312 respectively).
\begin{table}[!tbp]
\caption{Zero-shot Learning Accuracies. \label{zslres}}
\begin{center}
\footnotesize
    \begin{tabular}{|c|ccc|}
    \hline
     \multirow{2}*{Approach}
    &\multicolumn{3}{c|}{Classification Accuracies (\%)}\\ \cline{2-4}
&AWA&USAA&CUB\\
\hline
\textbf{HAP}&45.0&44.1$\pm$3.6&17.5\\
\textbf{CSHAP$_H$}&\textbf{45.6}&\textbf{45.3$\pm$4.2}&17.5\\
\textbf{CSHAP$_G$}&45.0&44.6$\pm$3.7&17.5\\
DAP~\cite{dap2014}&41.4/42.8$^*$&35.2&10.5\\
IAP~\cite{dap2014}&42.2/35.7$^*$&---&---\\
ALE~\cite{labelemb}&37.4&---&\textbf{18.0}\\
HLE~\cite{labelemb}&39.0&---&12.1\\
AHLE~\cite{labelemb}&43.5&---&17.0\\
M2LATM~\cite{lmla}&41.3&41.9&---\\
PST~\cite{tlts}&42.7&36.2&---\\
ZSRF~\cite{zsrf}&43.0&---&---\\
CLA~\cite{cla}&42.3&---&---\\
DA~\cite{decor}&30.6&---&---\\
      \hline
    \end{tabular}{}
    \end{center}
    \vspace{-0.8cm}
\end{table}

\vspace{-0.1cm}
\subsection{N-Shot Learning}
\begin{figure}[!tbp]
\setlength{\belowcaptionskip}{-0.4cm}
\centering
\subfigure[CUB]{
\centering
\includegraphics[scale=0.17]{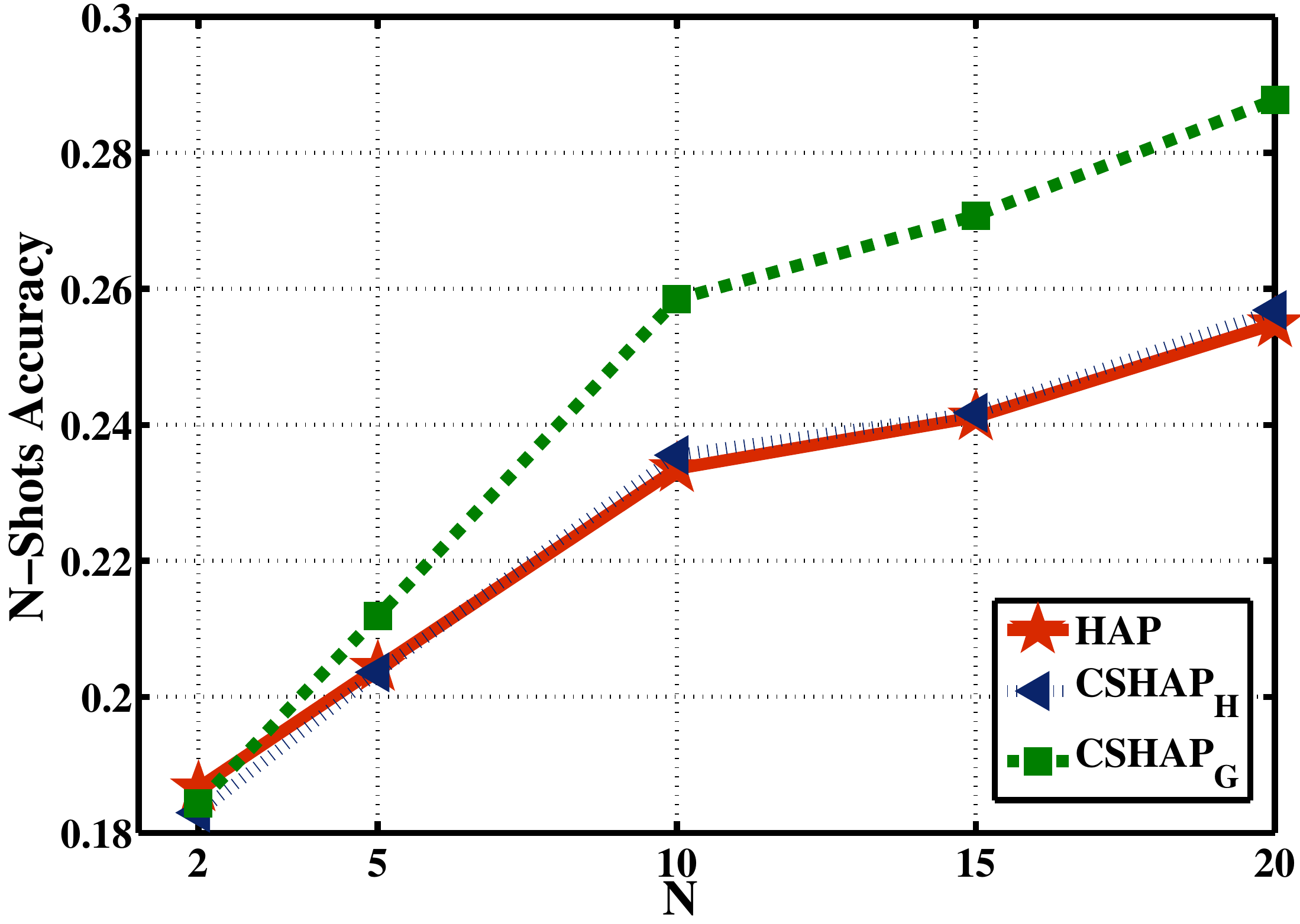}
}
\centering
\subfigure[USAA]{
\centering
\includegraphics[scale=0.17]{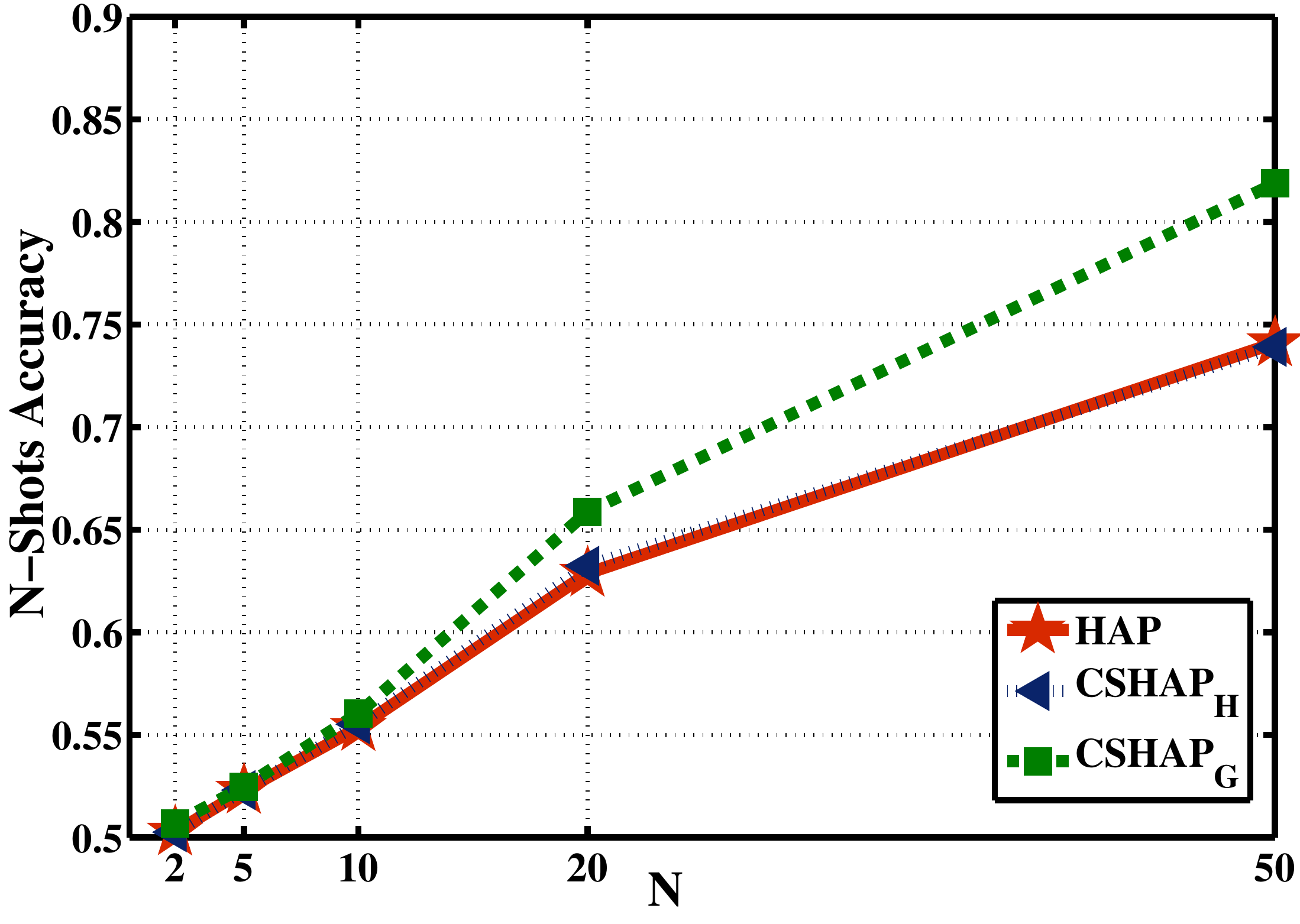}
}
\caption{N-Shot Learning accuracies on two datasets.}
\label{dims}
\vspace{-0.3cm}
\end{figure}

\vspace{-0.1cm}
We extend ZSL into N-Shot Learning (NSL) where a few ($N$) samples of the test classes are added to the training dataset. The experiments are conducted on the USAA and AWA datasets. Figure~\ref{dims} shows the trend of the three approaches as $N$ increases. In these experiments, we find that all three methods can get similar performances when $N$ is small. Interestingly, along increasing $N$, CSHAP$_G$  significantly outperforms others.  This confirms with our hypothesis about the way CSHAP$_G$ captures the intra-class manifold structure while incorporating the class information, in contrast to just grouping the homogenous samples together as CSHAP$_H$ does. In such case, the addition of the samples of the test classes improves the quality of the captured intra-class manifold structures, and therefore improves the performance of CSHAP$_G$.

Note the performances of HAP methods on USAA dataset are much better than the ones reported in section~\ref{zsl}, since only half of the testing samples and training samples are used, similar strategy is used in CUB dataset. Compared to the previous experiment $N$ samples per each test class have to be taken out for the NSL training.

\vspace{-0.1cm}
\subsection{Full Data Categorization}
\vspace{-0.1cm}
We consider the attributes as the feature representation to tackle the common categorization task. The default splits of data defined in the datasets are employed. Table~\ref{fdca} reports the results. Two classification methods are employed for classification. The first one is the one defined in Equation~\ref{ptempd}. The second one is the simple Euclidean distance-based Nearest Neighbor Classifier (NNC). The sign '$\dag$' indicates the results using NNC. Similar to the in NSL, CSHAP$_G$ achieves the best performances since it better integrates the class information.
\begin{table}[h]
\vspace{-0.1cm}
\caption{Full Data Categorization Accuracy (\%). \label{fdca}}
\footnotesize
\begin{center}
    \begin{tabular}{|c|cccc|}
    \hline
     \multirow{2}*{Database}
    &\multicolumn{4}{c|}{Classification Accuracies (\%)}\\ \cline{2-5}
&HAP&CSHAP$_H$&CSHAP$_G$&AHLE\\
      \hline
      CUB&23.1/21.4$^\dag$&23.1/21.2$^\dag$&\textbf{26.1}/24.5$^\dag$&23.5\\
      USAA&50.1/48.1$^\dag$&50.1/48.1$^\dag$&\textbf{50.9}/49.2$^\dag$&---\\
      \hline
    \end{tabular}{}
    \end{center}
    \vspace{-0.6cm}
\end{table}
\vspace{-0.1cm}
\subsection{Evaluations of Kernel HAP Algorithms}
\begin{figure}[!tbp]
\setlength{\belowcaptionskip}{-0.4cm}
\centering
\subfigure[CUB]{
\centering
\includegraphics[scale=0.175]{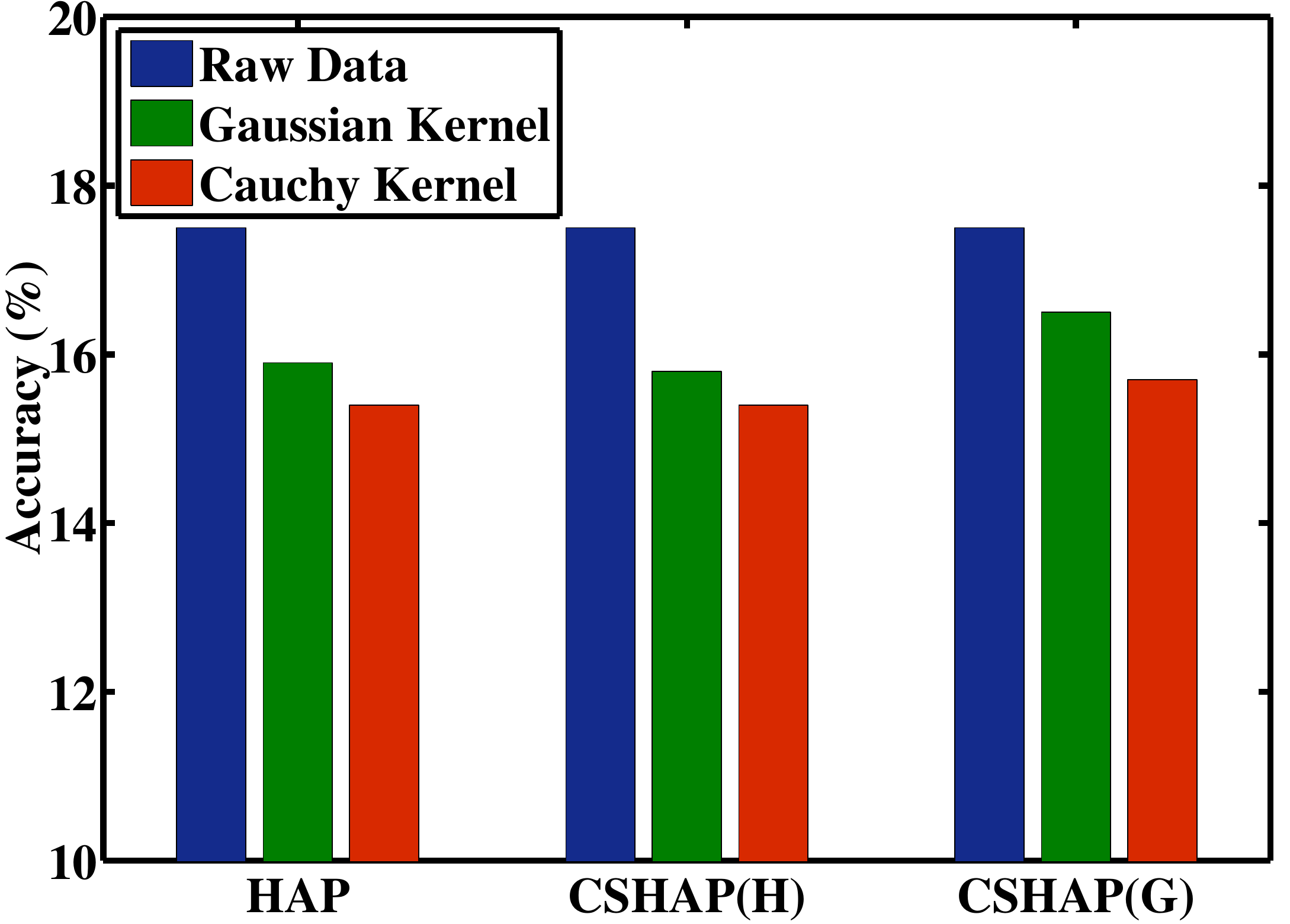}
}
\centering
\subfigure[USAA]{
\centering
\includegraphics[scale=0.175]{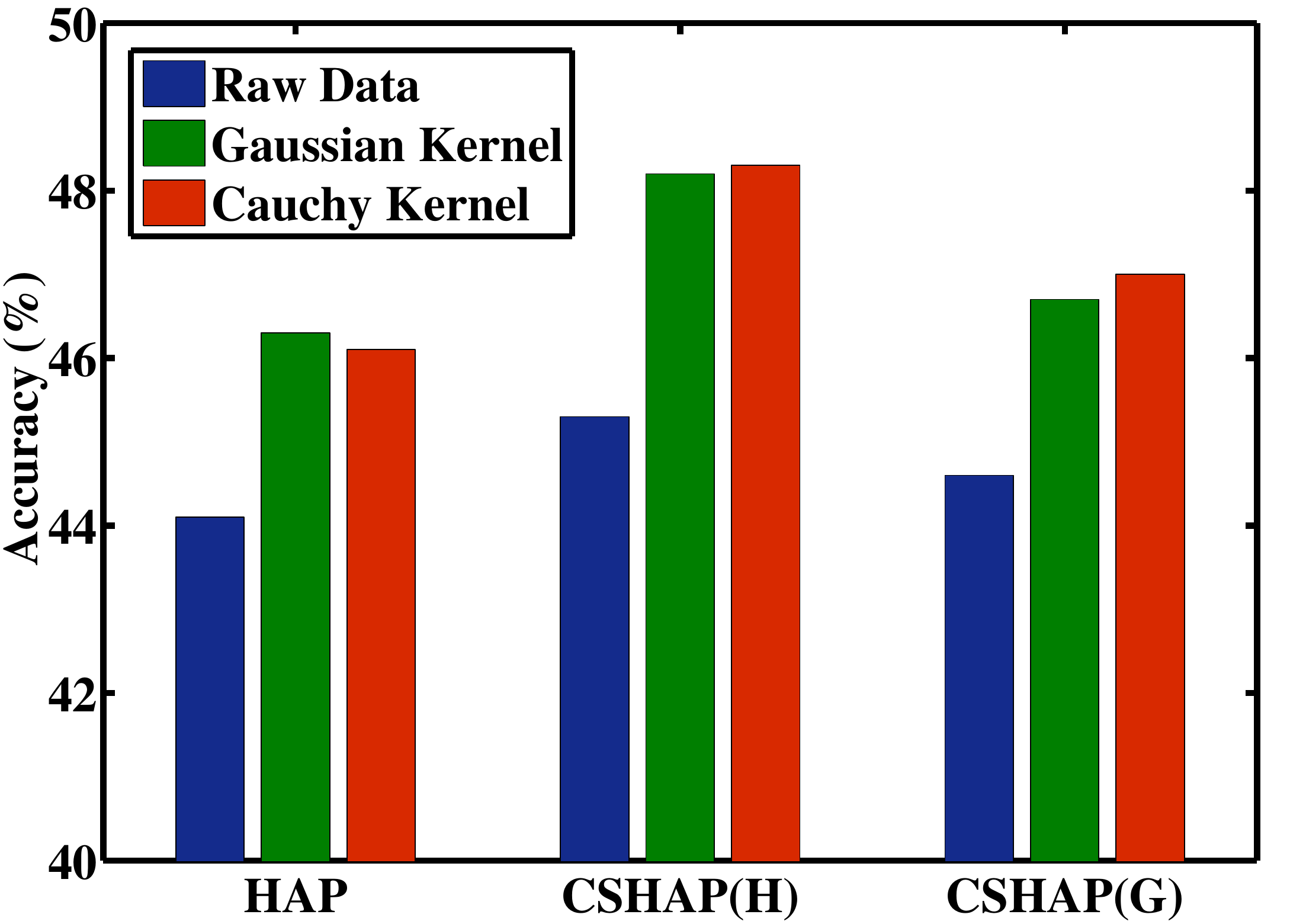}
}
\caption{The performances of Kernel HAP algorithms in USAA and CUB datasets.}
\label{kernelfig}
\vspace{-0.3cm}
\end{figure}

\vspace{-0.1cm}
We also conduct several experiments to test the potentials of the kernel HAP algorithms in Zero-Shot Learning. The Gaussian kernel and Cauchy kernel are applied.  Figure~\ref{kernelfig} shows the ZSL performances of these kernel HAP algorithms in USAA and CUB datasets. In USAA dataset, we can find that Gaussian kernel improves the ZSL accuracies of HAP, CSHAP$_H$ and CSHAP$_G$ from 44.1\%, 45.3\% and 44.6\% to 46.3\%, 48.2\% and 46.7\%. The ZSL accuracies of three Cauchy kernel-based HAP algorithms are 46.1\%, 48.3\% and 47.0\%. In CUB database, these two kernels actually reduced the ZSL accuracies of HAP algorithms. The accuracies of the kernel HAP algorithms is around 15.5\% to 16.5\%. We attribute this to the fact that the deep features used in CUB dataset are originally designed to be linear.

\vspace{-0.1cm}
\subsection{The Parameters and Computational Cost}
\vspace{-0.1cm}
The experimental results of the choices of the parameters are reported in the supplementary material. Since HAP is graph-based algorithm and involves the matrix inversion, its computational complexity for training is the minimum of $\mathcal{O}(nmd)$ and $\mathcal{O}(d^3)$ and its computational complexity of testing is $\mathcal{O}(d)$. So, it is more time consuming for training but quite efficient for testing. Taking the CUB dataset as an example (5994 samples for training and 5974 samples for testing), the time for training 312 attribute predictors is 23.33 seconds. The time for predicting the attributes of all test samples is 0.14 seconds. The code is written in matlab and the experimental hardware configuration is Quad-Core CPU: 2.5 GHz, RAM: 8G.

\vspace{-0.1cm}
\section{Conclusion}\label{s6}
\vspace{-0.1cm}
We presented a novel attribute prediction approach called Hypergraph-based Attribute Predictor (HAP) via deriving a collection of attribute classifiers from the hypergraph embedding, in which the attribute relations are considered as hyperedges and the hypergraph cuts are the attribute predictions. The hypergraph formulation facilitates exploiting the correlations of the attributes as well as jointly learning the attribute predictors. Moreover, the additional information can be flexibly incorporated into HAP via encoding the information in a penalty graph or hypergraph. To generalize the mappings between the feature space and attribute space which are known as the attribute predictors, we also kernelized the model. Extensive experiments on three well known attribute datasets demonstrated the effectiveness of our model for attribute prediction, Zero-Shot Learning, N-Shot Learning and categorization. From the results we can conclude that the CSHAP$_G$ variant is the best to integrate class labels for N-shot learning, however the three proposed variants performs similarly in zero-shot learning task.

\section*{Acknowledgement}
The work described in this paper was partially supported by the National Natural Science Foundation of China (Grant no. 61173131,91118005, 11202249), Program for Changjiang Scholars and Innovative Research Team in University (Grant No. IRT1196) and the Fundamental Research Funds for the Central Universities (Grant Nos. CDJZR12098801 and CDJZR11095501). Dan Yang is the corresponding author of this paper.

\vspace{-0.2cm}
{\footnotesize
\bibliographystyle{ieee}
\bibliography{mybib}
}
{\onecolumn 
\newpage
\begin{center}
{\huge
\textbf{The Supplementary Materials}}
\end{center}

\section*{The Derivation Details of Attribute Relation Loss Function}
 In this section, we will introduce the detailed derivations of Equation (6). Before it, let us review some related notations. The attribute relations are encoded in a given hypergraph $G=(V,E)$ where $V$ is the vertex set and $E$ denotes the hyperedge sets. In this hypegraph, each vertex is corresponding to an instance and each hyperedge is associated with an attribute relation. The degree of a hyperedge $e \in E$, which is denoted as $\delta(e)$, is the number of vertices in $e$. The $(v,e)$-th element of vertex-edge incidence matrix $H\in \mathcal{R}^{|V|\times |E|}$ is considered as $h(v,e)=1$ if $v \in e$ otherwise $h(v,e)=0$ where $v\in V$. $d(v)=\sum_{v\in e, e\in E} w(e)=\sum_{e\in E} w(e)h(v,e)$ denotes the degree of the vertex $v$. $w(e)$ is the weight of the hyperedge $e$. We denote the diagonal matrix forms of $\delta(e)$, $d(v)$ and $w(e)$ as $D_e$, $D_v$ and $W$ respectively. We obtain the attribute predictions by defining a collection of hypergraph cuts which is denoted as $F$. $F_u$ returns a row vector of $F$ which is corresponding to the predictions of attributes for the vertex $u$. $L_H$ is the normalized hypergraph Laplacian matrix which is derived from the hypergraph of attributes, and $I$ is an identity matrix. $\text{Tr}(\cdot)$ is the trace of the matrix.

Now we can present the detailed derivations of Equation (6) as follows:

\begin{eqnarray}\label{infloss}\nonumber
  \Omega(F, G)&=&\frac{1}{2}\sum_{e\in E}\sum_{(u,v)\in e}\frac{w(e)}{\delta(e)}\left|\left|\frac{F_u}{\sqrt{d(u)}}-\frac{F_v}{\sqrt{d(v)}}\right|\right|^2\\ \nonumber
  &=&\sum_{e\in{E}}\sum_{u,v\in V}\frac{w(e)h(u,e)h(v,e)}{\delta{(e)}}(\frac{(F_u)^2}{d(u)}-\frac{F_uF_v^T}{\sqrt{d(u)d(v)}})\\ \nonumber
&=&\sum_{e\in{E}}\sum_{u\in V}\frac{w(e)h(u,e)(F_u)^2}{d(u)}\sum_{v\in V}\frac{h(v,e)}{\delta(e)}
-\sum_{e\in{E}}\sum_{u,v\in V}\frac{F_uw(e)h(u,e)h(v,e)F_v^T}{\delta{(e)}\sqrt{d(u)d(v)}}\\
&=&\sum_{e\in{E}}(F_u)^2\sum_{u\in V}\frac{w(e)h(u,e)}{d(u)}
-\sum_{e\in{E}}\sum_{u,v\in V}\frac{F_uw(e)h(u,e)h(v,e)F_v^T}{\delta{(e)}\sqrt{d(u)d(v)}}\\ \nonumber
&=&\sum_{e\in{E}}(F_u)^2-\sum_{e\in{E}}\sum_{u,v\in V}\frac{F_uw(e)h(u,e)h(v,e)F_v^T}{\delta{(e)}\sqrt{d(u)d(v)}}\\ \nonumber
  &=&\text{Tr}(F^TF)-\text{Tr}(F^TD_v^{-1/2}{HW D_e^{-1}H^T}{D_v^{-1/2}}F)\\\nonumber
  &=&\text{Tr}(F^T(I-D_v^{-1/2}{HW D_e^{-1}H^T}{D_v^{-1/2}})F)\\ \nonumber
  &=&\text{Tr}(F^TL_HF),
\end{eqnarray}

\section*{The Influences of Parameters}
There are several parameters in HAP models. They are $\mu$, $\lambda$, $\eta$ and $\gamma$. $\mu$ is used for controlling the degree of hyperedge weighting. $\lambda$ is used for controlling the trade off between the attribute relation loss and the attribute prediction error. $\eta$ is employed for avoiding the overfitting. $\gamma$ is adopted for controlling the degree of penalty of the side information loss. The ultimate goals of different attribute learning-based systems are different. Some systems may aim at annotation or retrieval. These systems pay more attention on the improvement of the attribute prediction accuracy. Some other systems may focus on the categorization, \ie, Zero-shot Learning, N-shot Learning and Attribute-based categorization. These systems pay more attention on the exploitation of discriminating power of attributes. In our approach, it is available for us to tune the parameters to decide which evaluation metric we care more. Therefore, we will separately discuss the choices of parameters in these two cases.
\begin{figure*}[!tbp]
\setlength{\belowcaptionskip}{-0.4cm}
\centering
\subfigure[AWA]{
\centering
\includegraphics[scale=0.25]{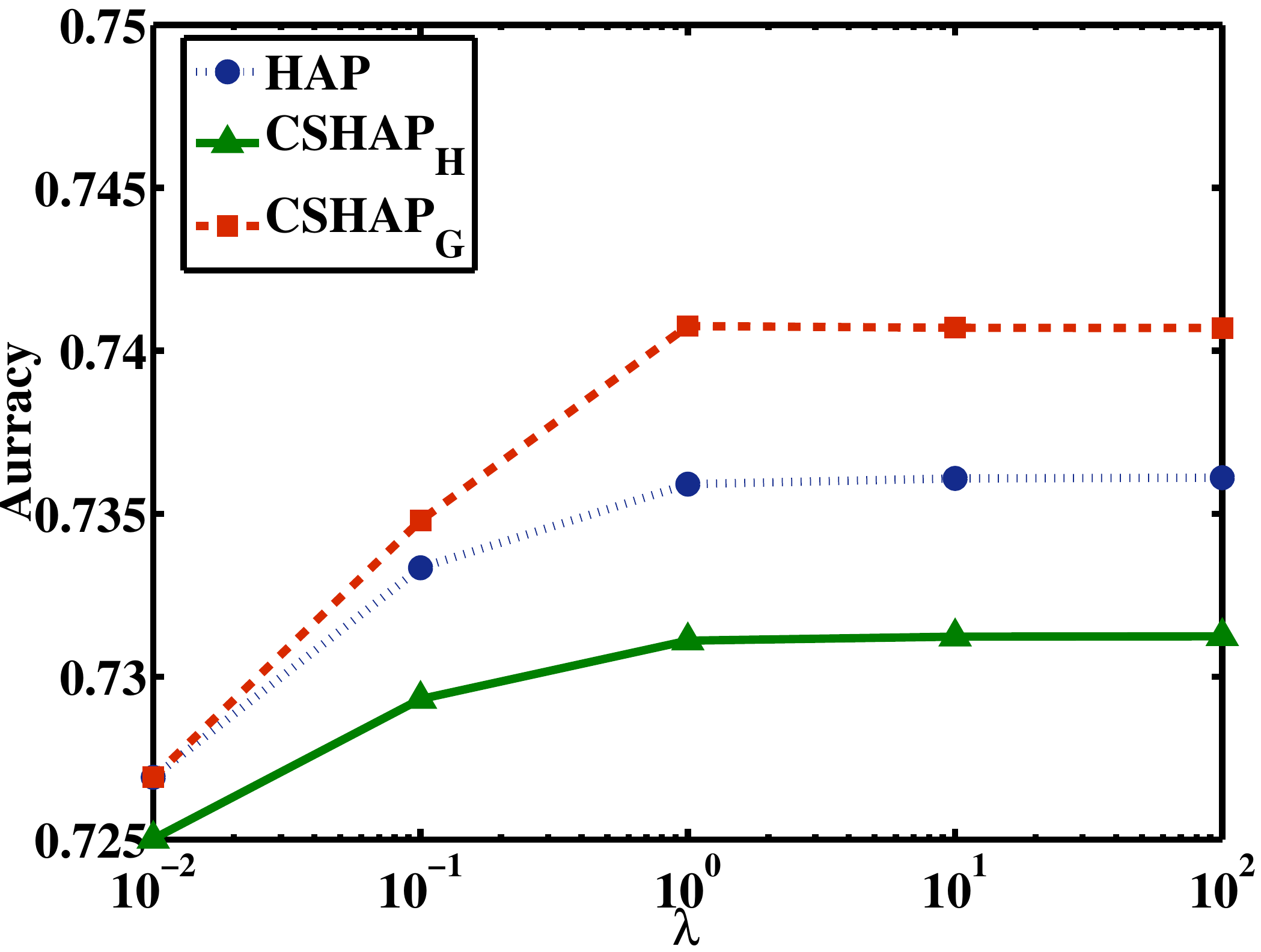}
}
\centering
\subfigure[USAA]{
\centering
\includegraphics[scale=0.25]{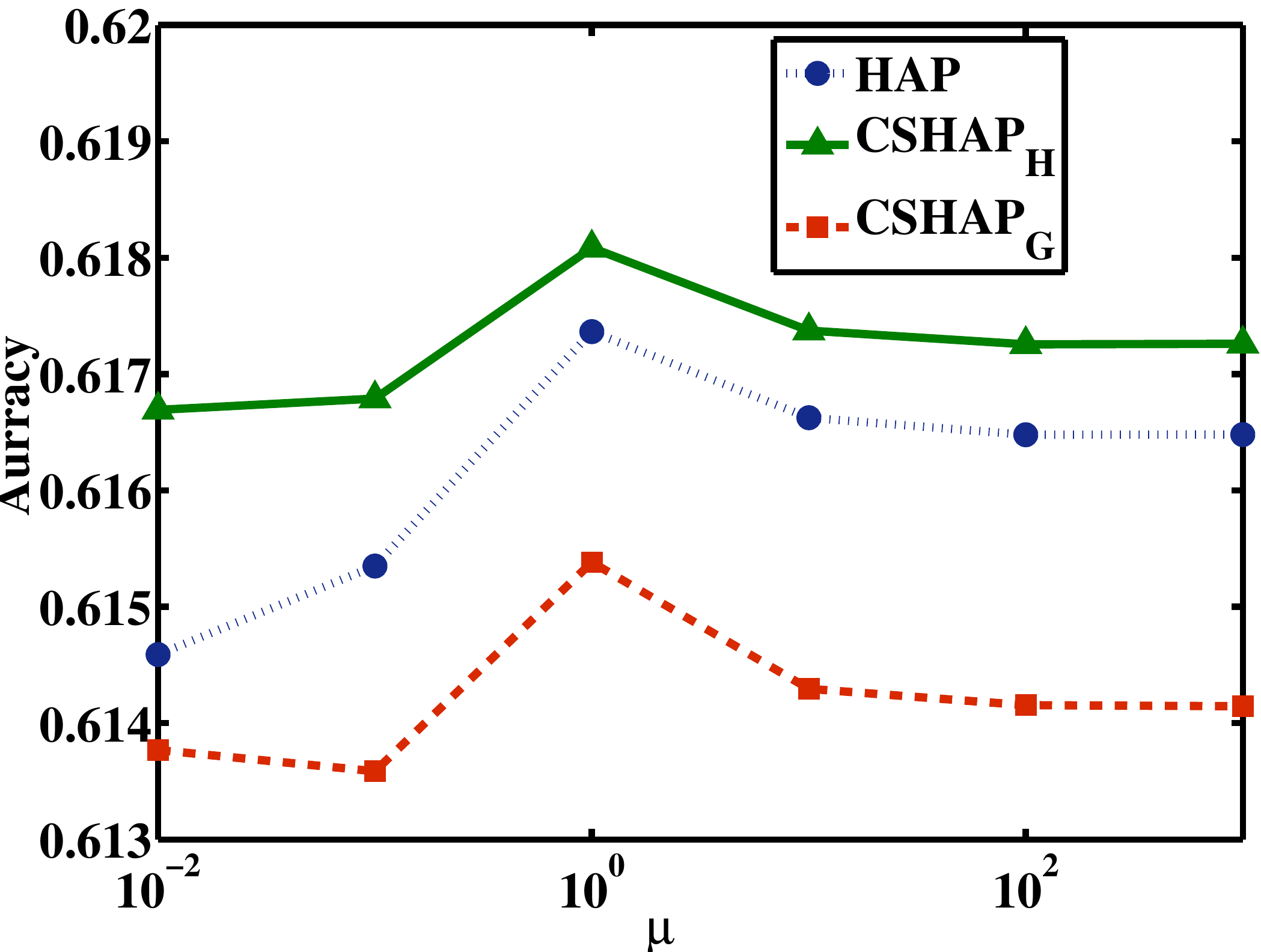}
}
\centering
\subfigure[CUB]{
\centering
\includegraphics[scale=0.25]{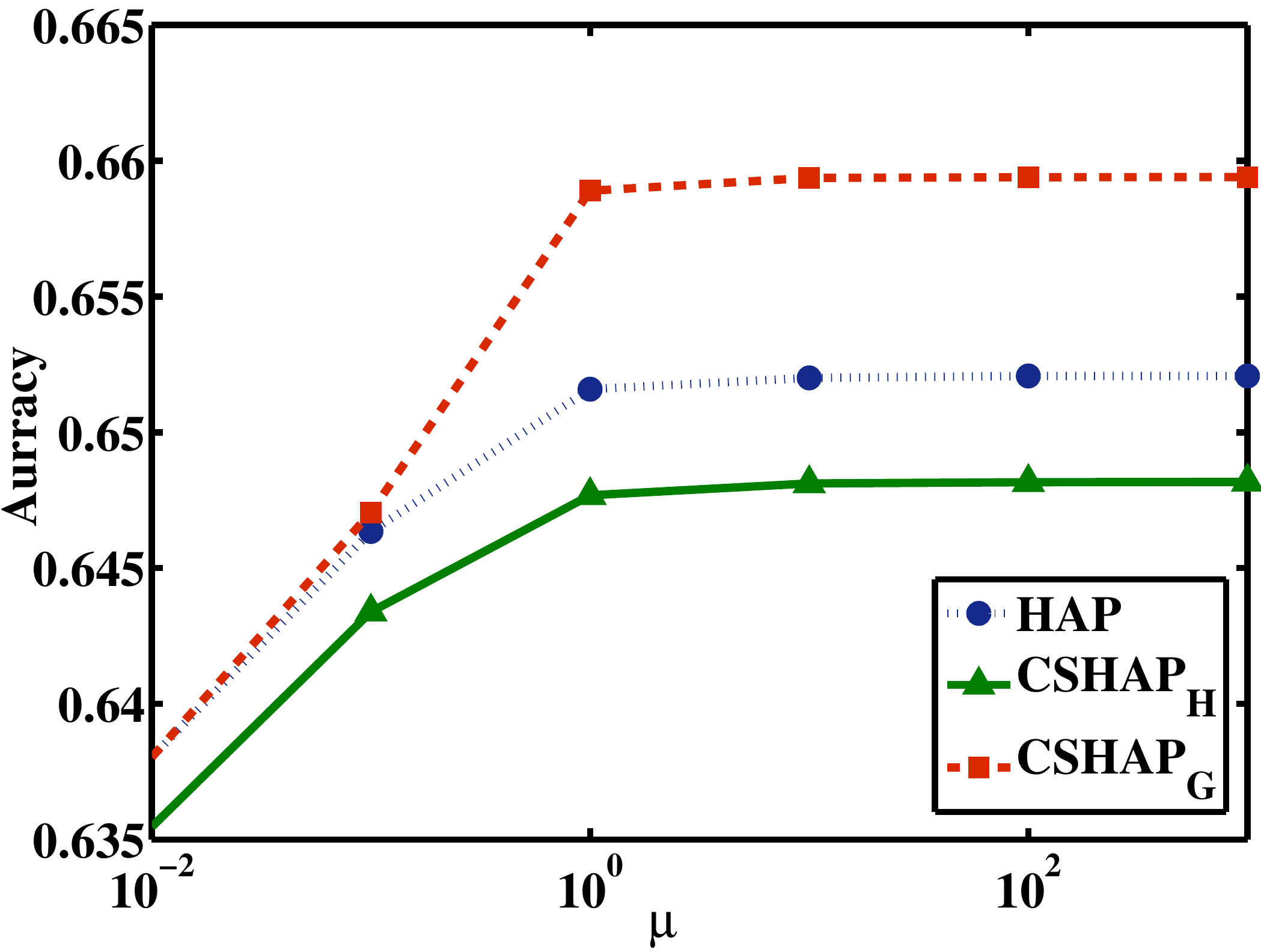}
}
\caption{The influences of $\mu$ to the attribute prediction accuracies.}
\label{muauc}
\end{figure*}
\begin{figure*}[!tbp]
\setlength{\belowcaptionskip}{-0.4cm}
\centering
\subfigure[AWA]{
\centering
\includegraphics[scale=0.25]{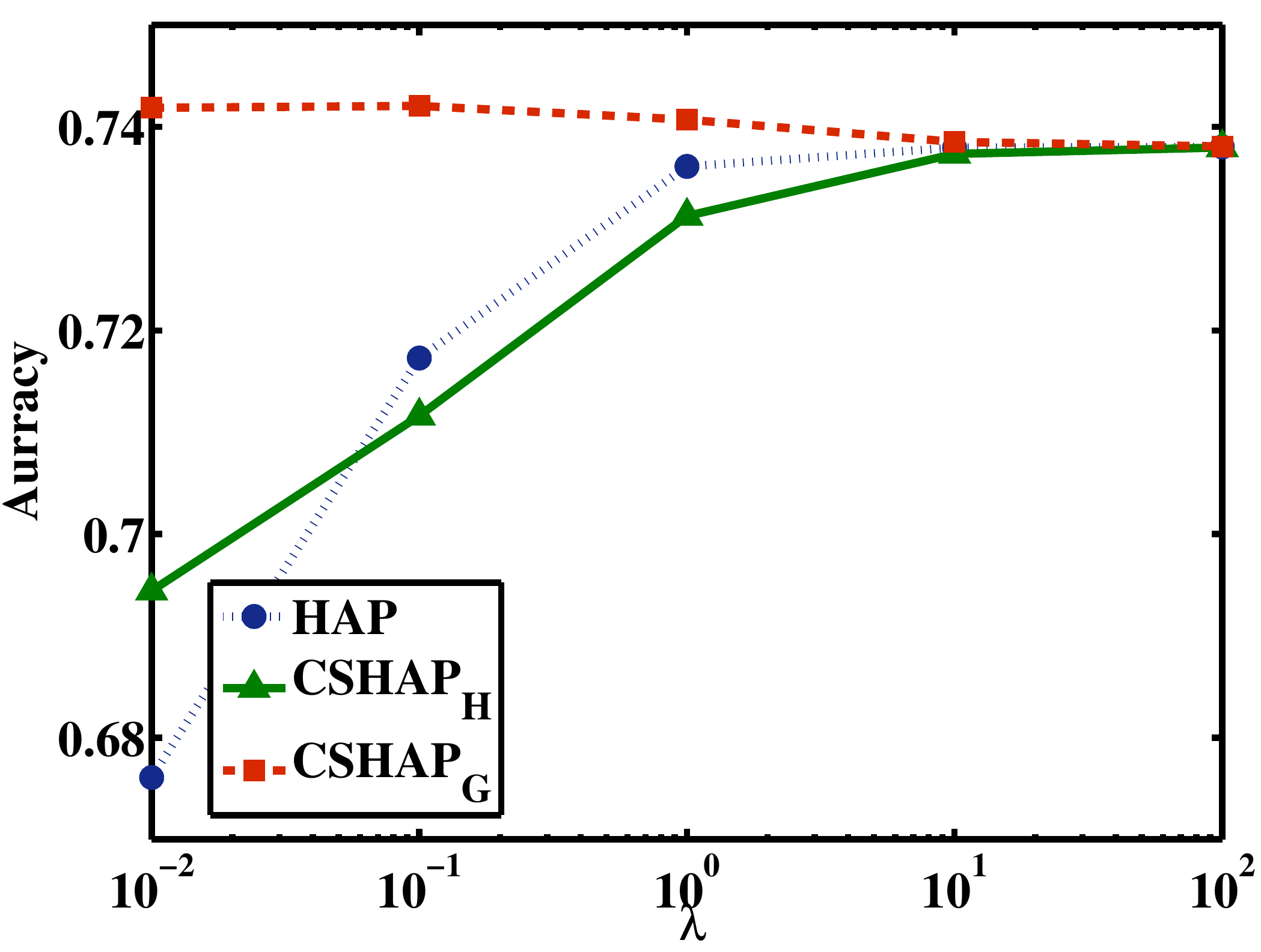}
}
\centering
\subfigure[USAA]{
\centering
\includegraphics[scale=0.25]{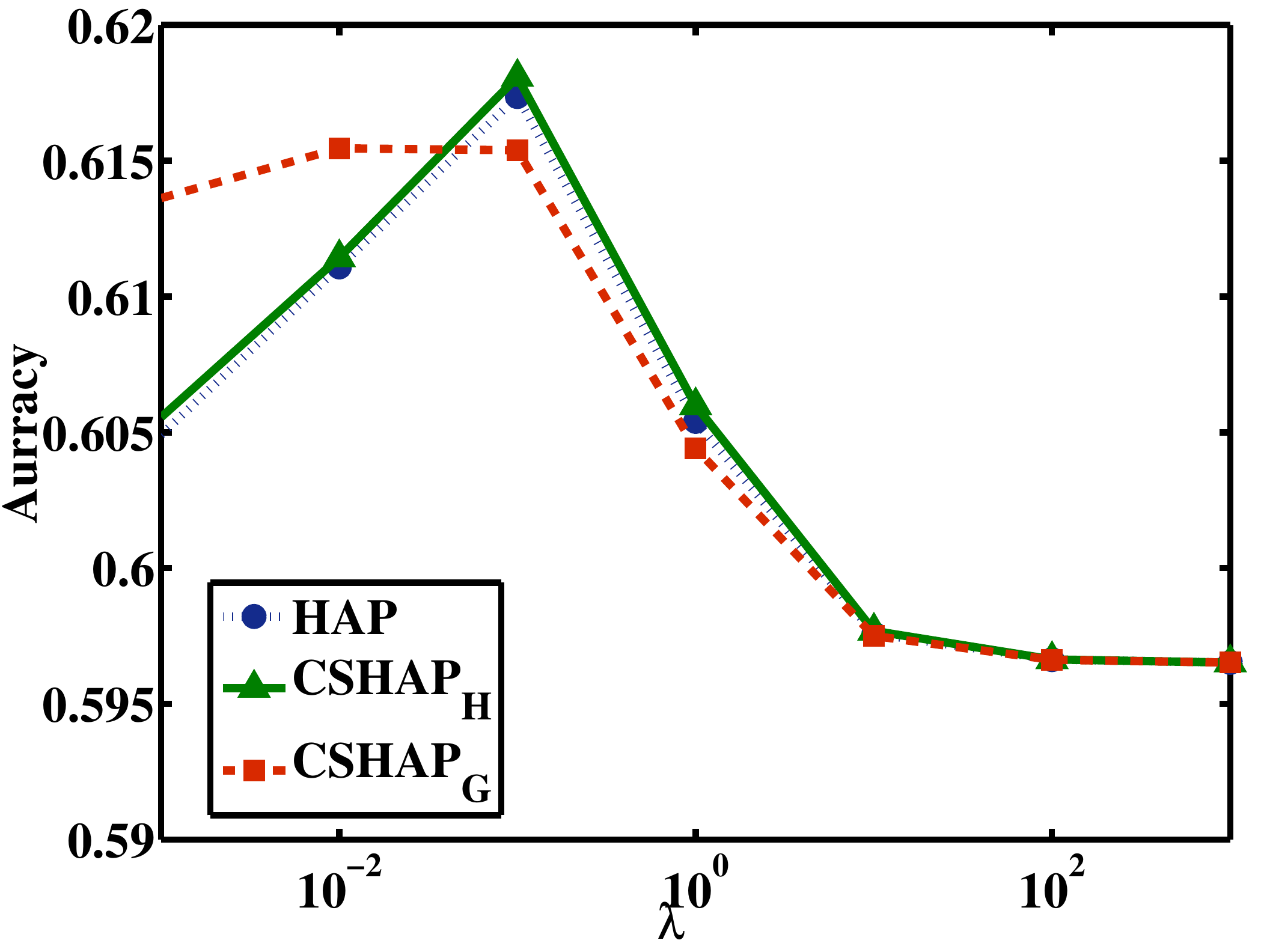}
}
\centering
\subfigure[CUB]{
\centering
\includegraphics[scale=0.25]{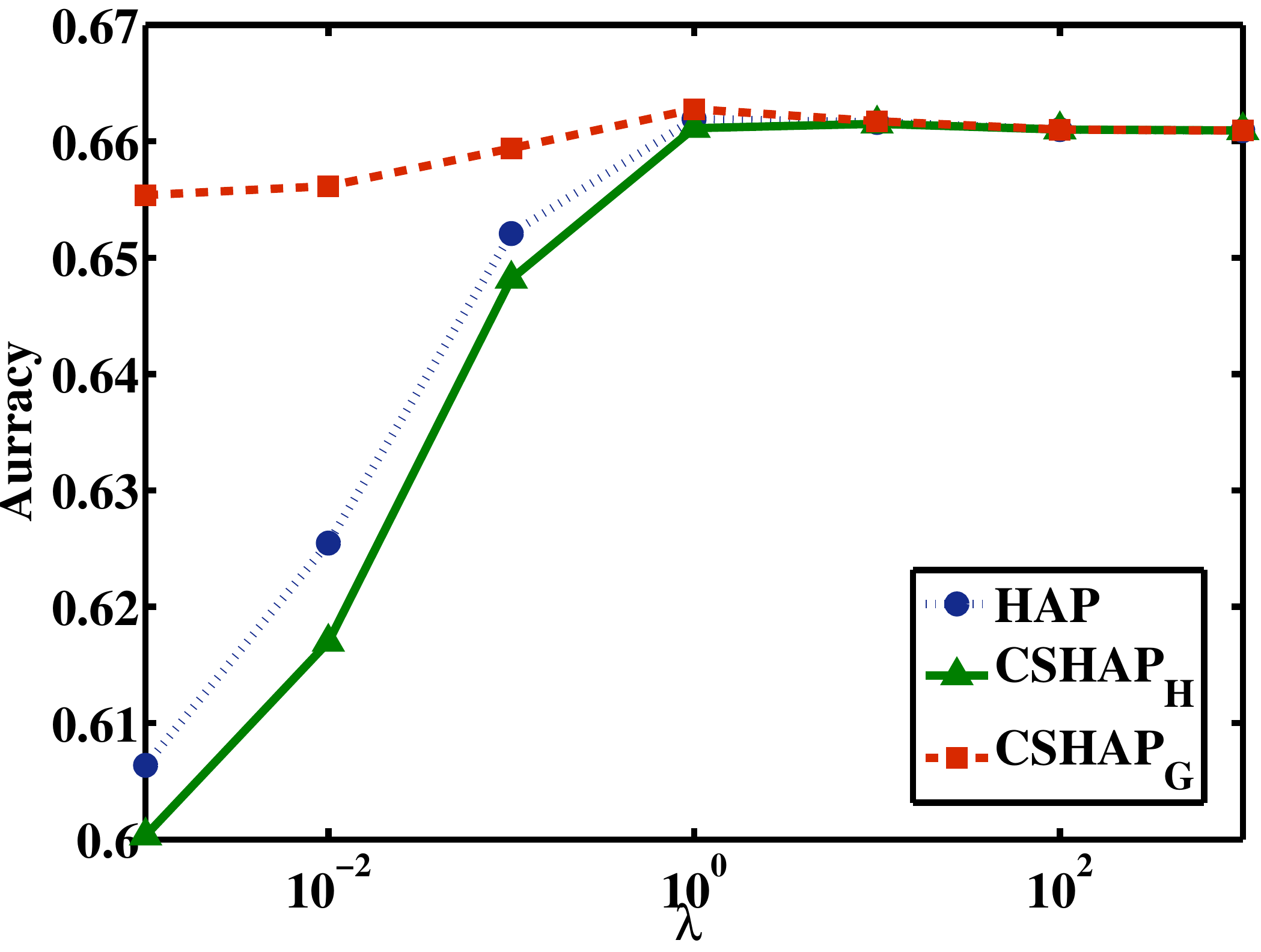}
}
\caption{The influences of $\lambda$ to the attribute prediction accuracies.}
\label{lambdaauc}
\end{figure*}
\begin{figure*}[!tbp]
\setlength{\belowcaptionskip}{-0.4cm}
\centering
\subfigure[AWA]{
\centering
\includegraphics[scale=0.25]{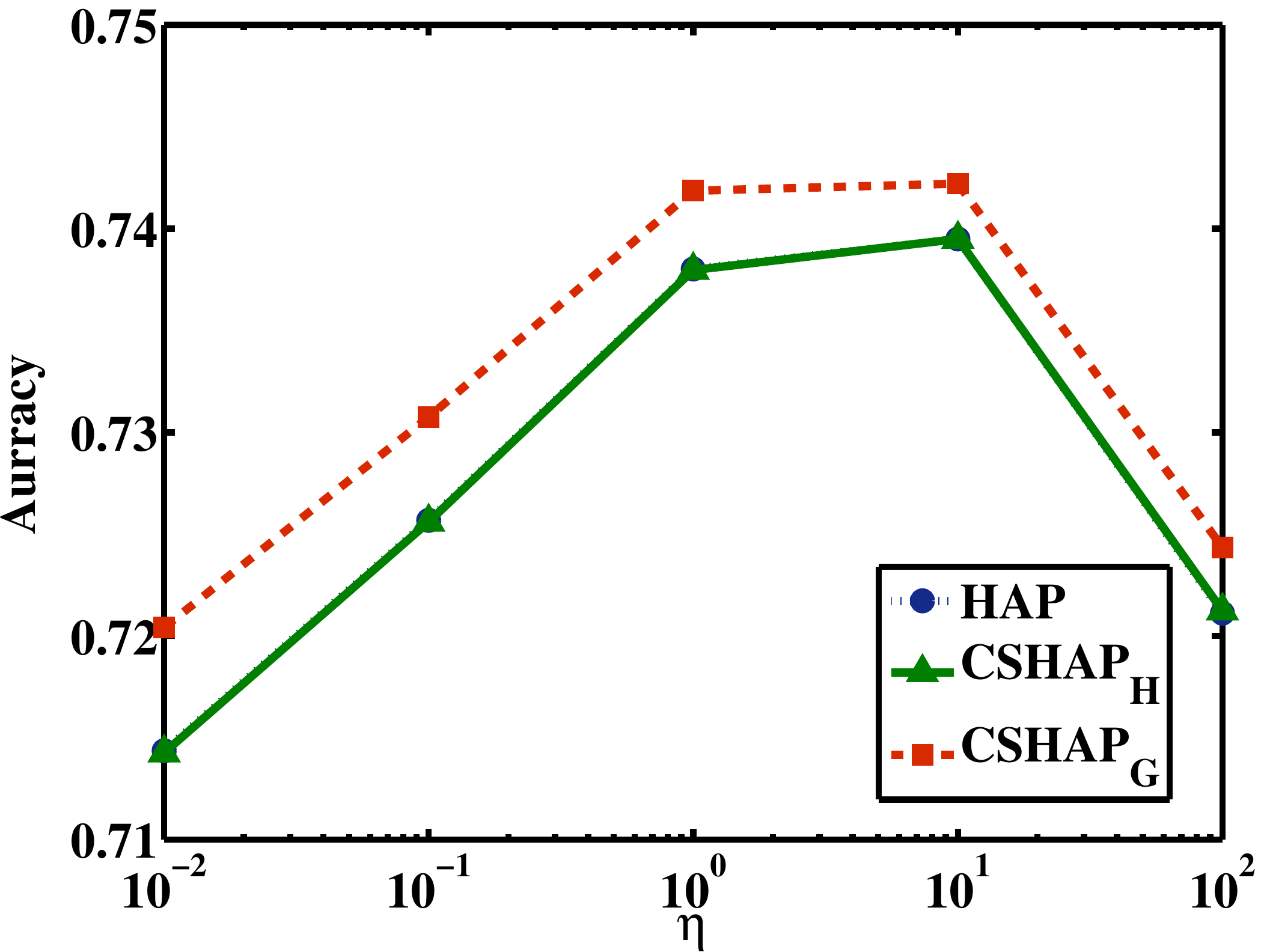}
}
\centering
\subfigure[USAA]{
\centering
\includegraphics[scale=0.25]{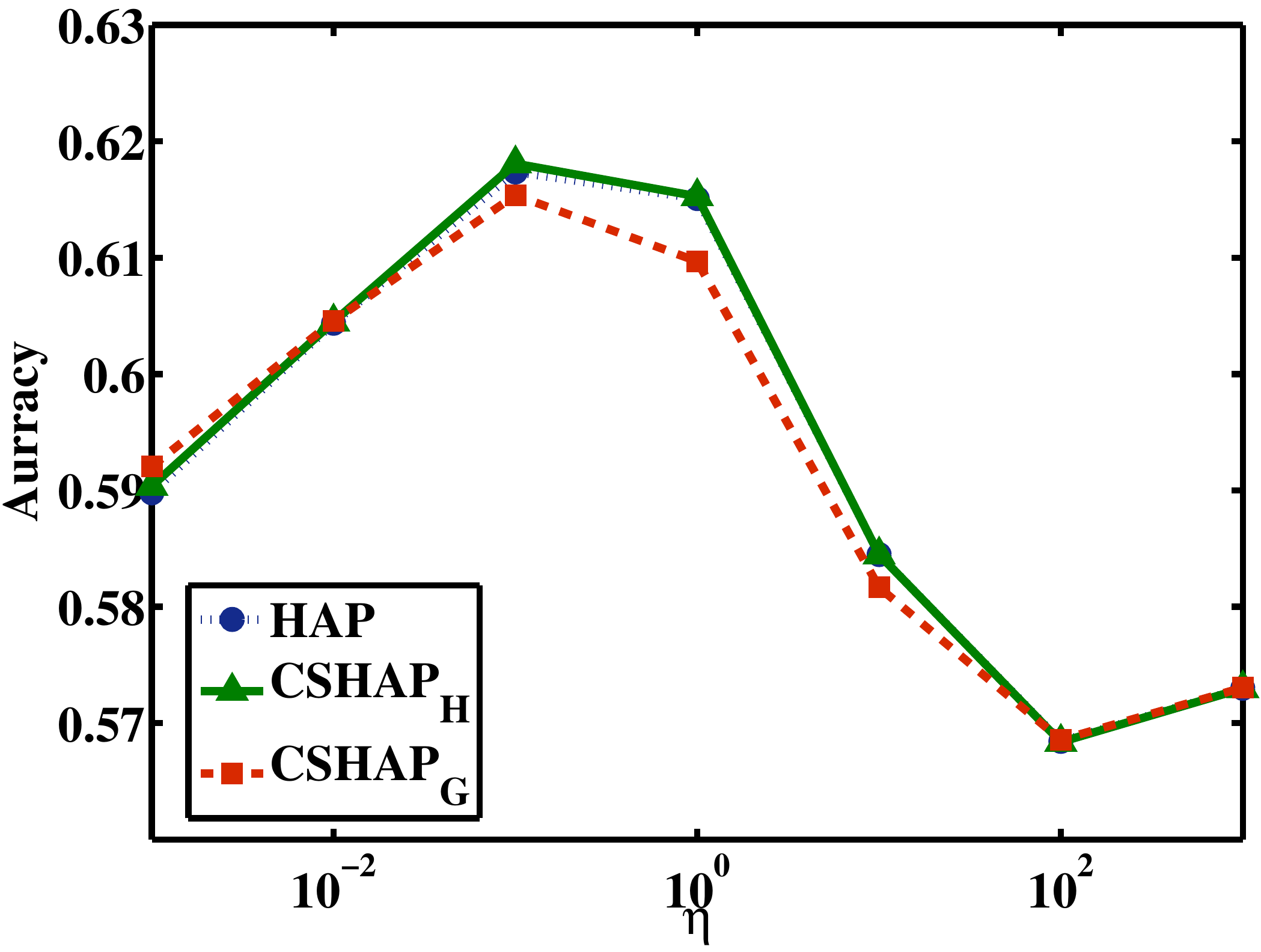}
}
\centering
\subfigure[CUB]{
\centering
\includegraphics[scale=0.25]{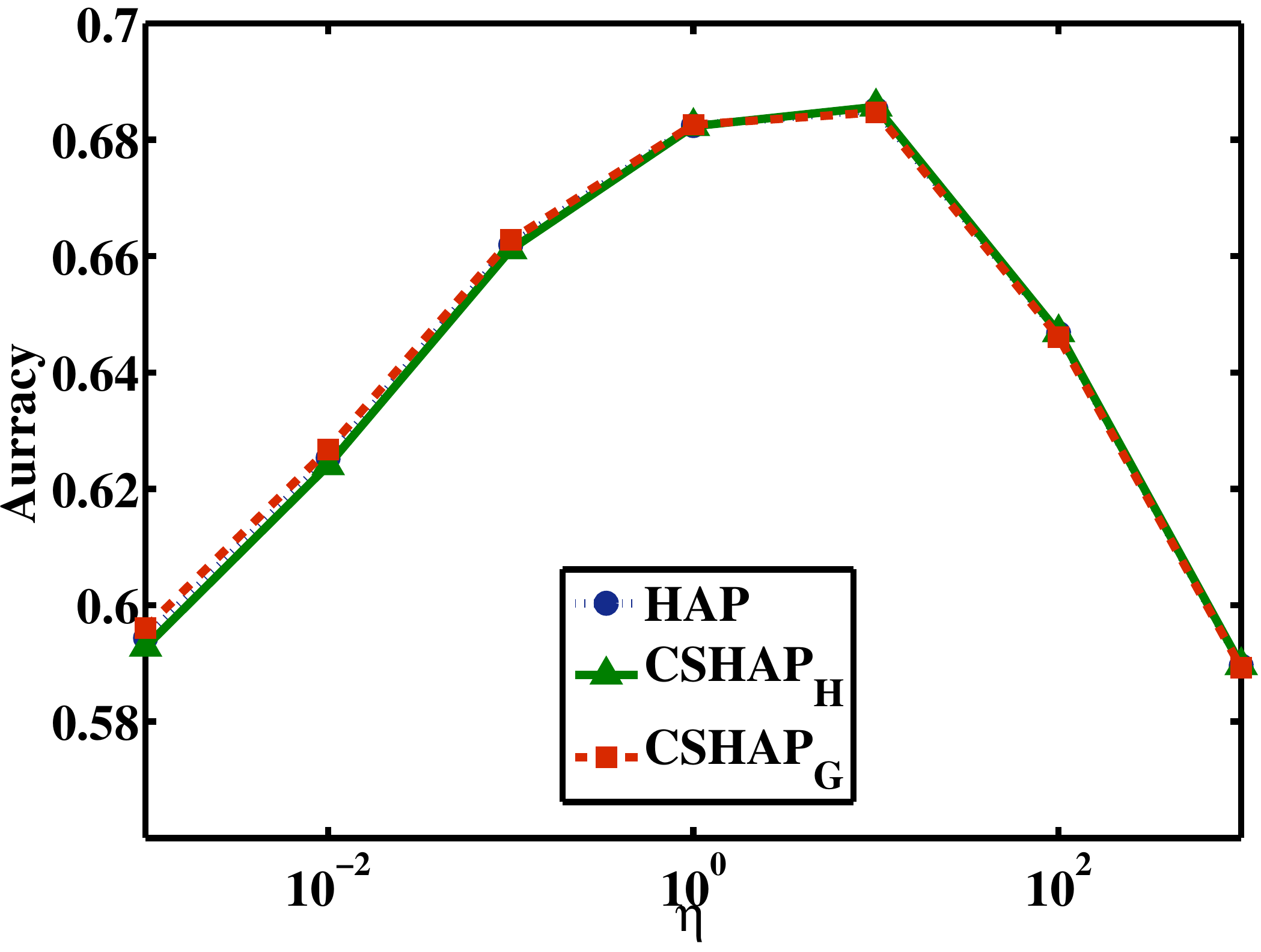}
}
\caption{The influences of $\eta$ to the attribute prediction accuracies.}
\label{etaauc}
\end{figure*}
\begin{figure*}[!h]
\setlength{\belowcaptionskip}{-0.4cm}
\centering
\subfigure[AWA]{
\centering
\includegraphics[scale=0.25]{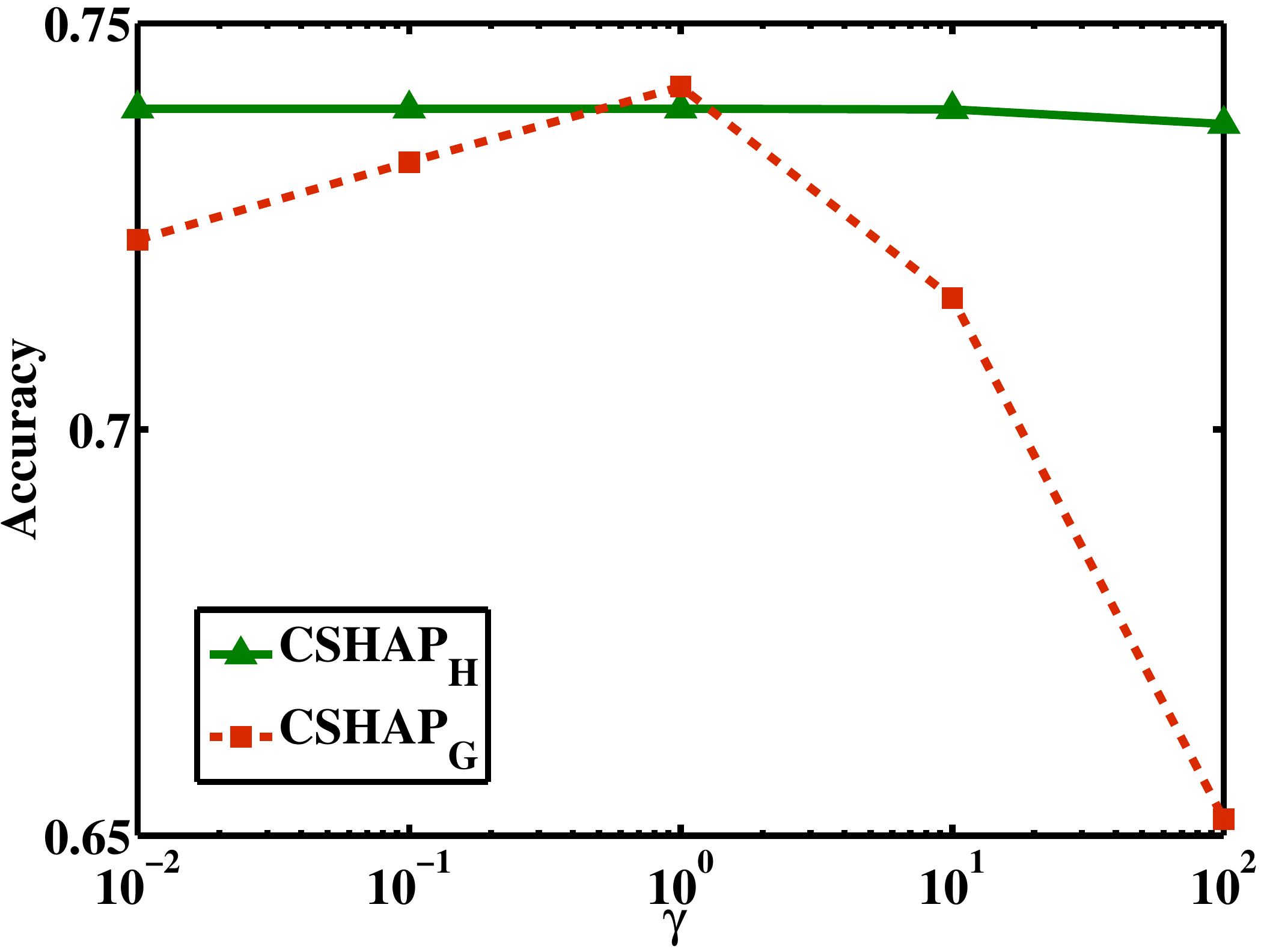}
}
\centering
\subfigure[USAA]{
\centering
\includegraphics[scale=0.25]{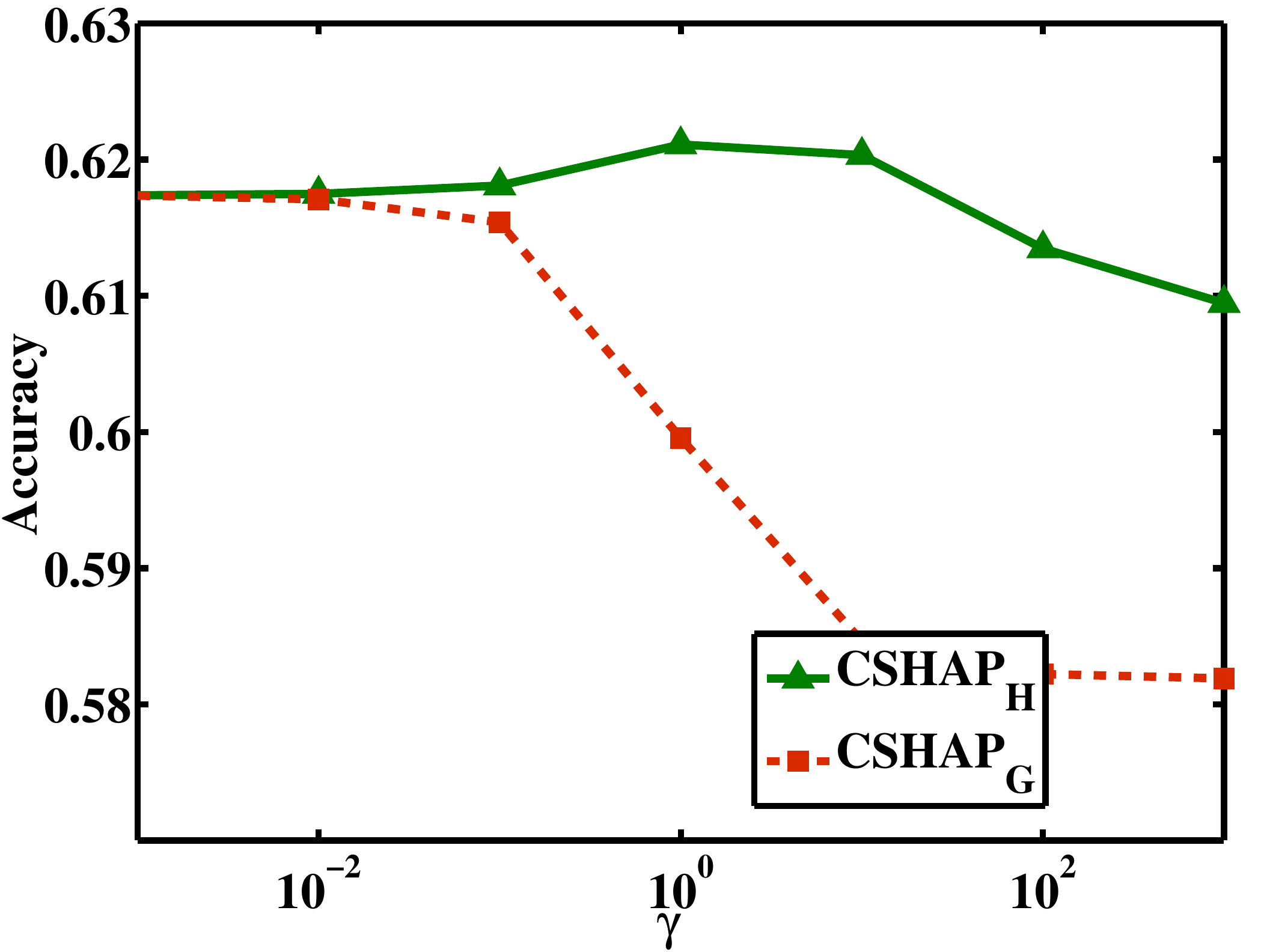}
}
\centering
\subfigure[CUB]{
\centering
\includegraphics[scale=0.25]{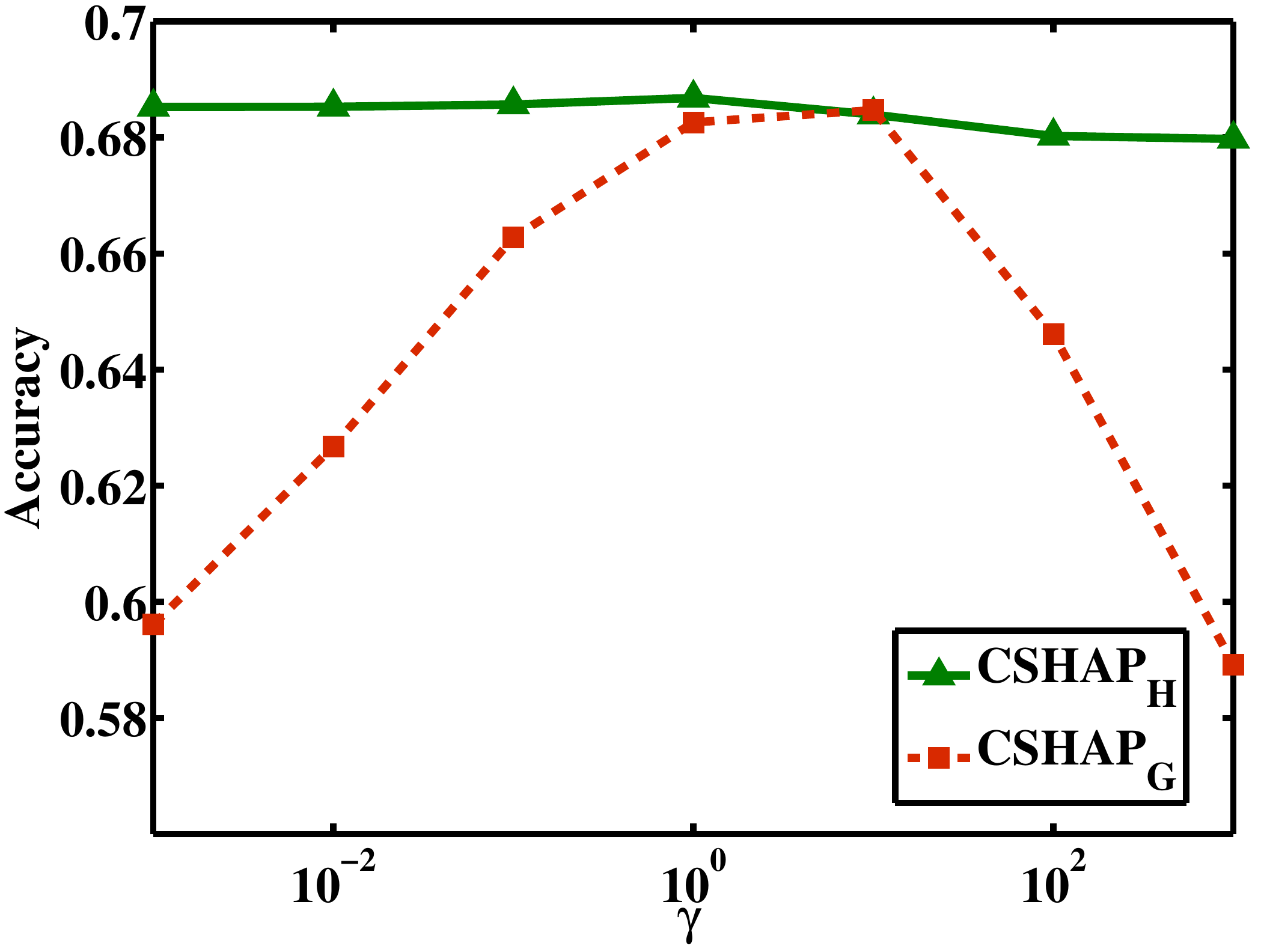}
}
\caption{The influences of $\gamma$ to the attribute prediction accuracies.}
\label{gammaauc}
\end{figure*}

\subsection*{The Influences of Parameters to The Attribute Prediction}
HAP has three parameters, $\mu$, $\lambda$ and $\eta$, need to tuned. CSHAP algorithms have one more parameter $\gamma$ need to be tuned. In the parameter selection procedure, we choose one parameter to tune and fix the values of the other parameters. The initial values of $\mu$, $\lambda$, $\eta$ and $\gamma$ are equal to 0.1. The parameter selection procedure is start from $\mu$ to $\gamma$. Once the optimal value of a parameter is learned, its corresponding initial value is replaced by that optimal value for more accurately estimating the optimal values of the rest parameters. Figures~\ref{muauc},~\ref{lambdaauc} and~\ref{etaauc}, respectively reports the attribute prediction accuracies using different $\mu$, $\lambda$, $\eta$ on different databases. From the observations, all HAP algorithms can achieve the best performances on all three databases when $\mu = 1$; The best choices of $\lambda$ for AWA, USSA and CUB databases are 10, 0.1 and 1 respectively and such numbers of $\eta$ are 10, 0.1 and 10. Figure~\ref{gammaauc} plots the relationships between $\gamma$ and the attribute prediction accuracy on three different databases. We can find that CSHAP$_G$ is more sensitive to $\gamma$. It is not hard to conclude from the observations that the optimal values of $\gamma$ are 1, 0.01 and 10 for AWA, USAA and CUB databases respectively.

\subsection*{The Influences of Parameters to Zero-Shot Learning}
In Zero-Shot Learning (ZSL), we need to employ the sigmoid function to normalize the attribute confidences, which are obtained by our models, into range [0,1]. So, there is one additional parameter $\rho$ should be studied in this section. We follow the aforementioned parameter selection manner to select the parameters. The selection procedure is start from $\mu$ to $\rho$ where the initial value of $\rho$ is 0.5. Figure~\ref{mu} shows the ZSL accuracies under different $\mu$. On AWA and CUB database, all three approach can get the best performances when $\mu=0.1$ while the optimal value of $\mu$ on USAA database is 1. Compared with other parameters, HAP algorithms are relatively insensitive to $\mu$ when its value is bigger than 1. Figures~\ref{lambda} and~\ref{eta} demonstrate the impacts of ZSL accuracies from $\lambda$ and $\eta$ respectively. The curves of these figures share similar behavior that their peaks are very explicit. From the observations, we can know that the optimal values of $\lambda$ are 1, 0.01 and 0.1 on AWA, USAA and CUB databases respectively while such numbers of $\eta$ are 1, 0.1 and 0.1. As same as the phenomenon observed in Figure~\ref{gammaauc}, Figure~\ref{gamma} also shows that CSHAP$_G$ is very sensitive to $\gamma$ but CSHAP$_H$ is robust to $\gamma$. Here, we suggest to set the $\gamma$ of AWA, USAA and CUB databases to 1, $10^{-3}$ and $10^{-3}$ respectively. Figure~\ref{rho} reports the ZSL performances under different $\rho$. However, it is really hard to conclude uniform setting for each database. So we choose different $\rho$ for different approaches. More specifically, we suggest to choose the $\rho$ in the range [0.007, 0.02] for CSHAP$_H$ while choose the $\rho$ in the range [0.06,0.1] for HAP and CSHAP$_G$ on AWA database. On USAA database, CSHAP$_G$ can get good ZSL performances when $\rho$ is in the range [0.005,0.01] while the good $\rho$ for CSHAP$_H$ and HAP should be above 0.3. The impacts of $\rho$ to the performances of all three algorithms are similar on CUB databases. The observations indicate that the $\rho$ which is larger than 1 can get the good performances for all three HAP algorithms.
\begin{figure*}[h]
\setlength{\belowcaptionskip}{-0.4cm}
\centering
\subfigure[AWA]{
\centering
\includegraphics[scale=0.25]{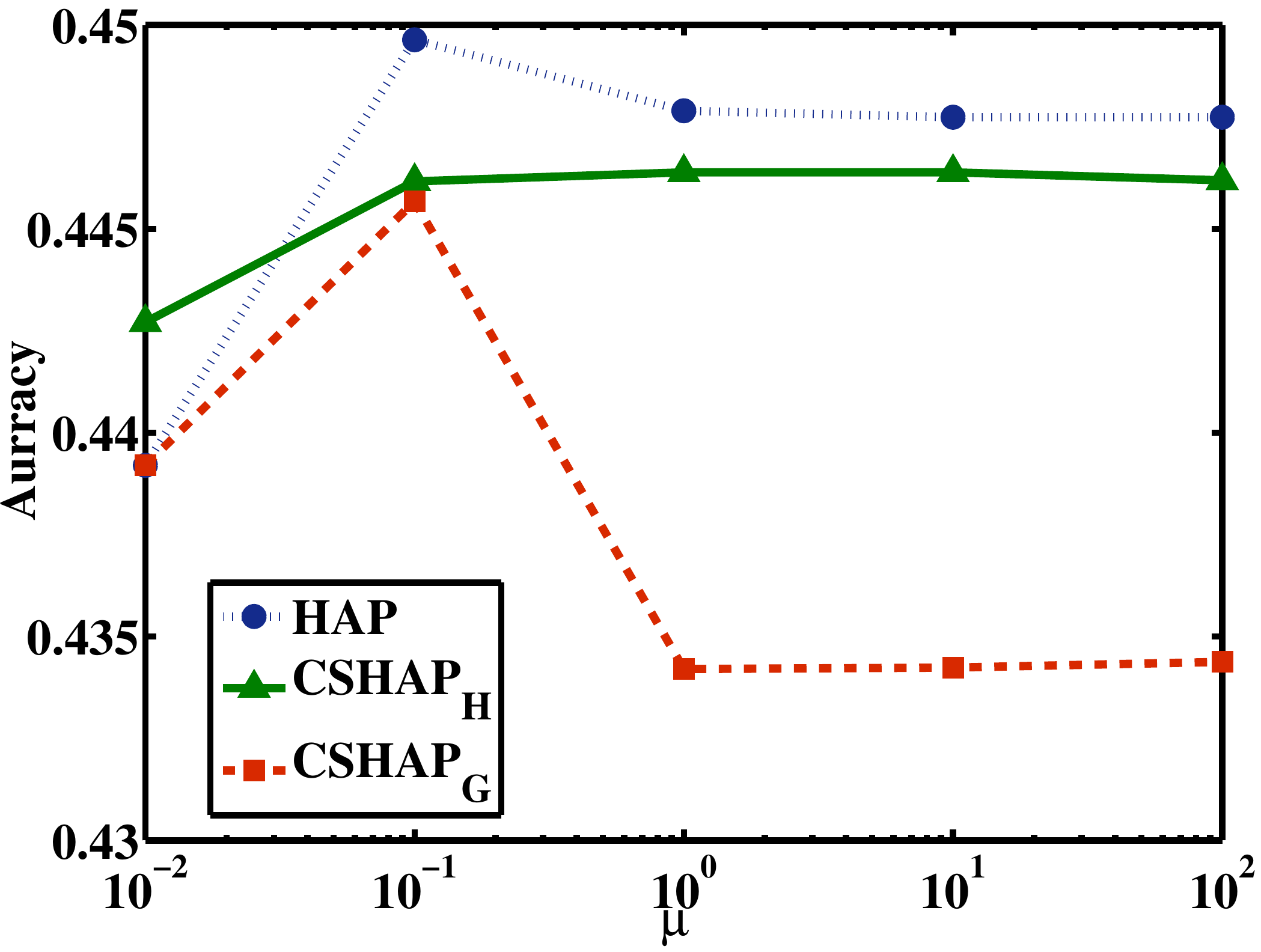}
}
\centering
\subfigure[USAA]{
\centering
\includegraphics[scale=0.25]{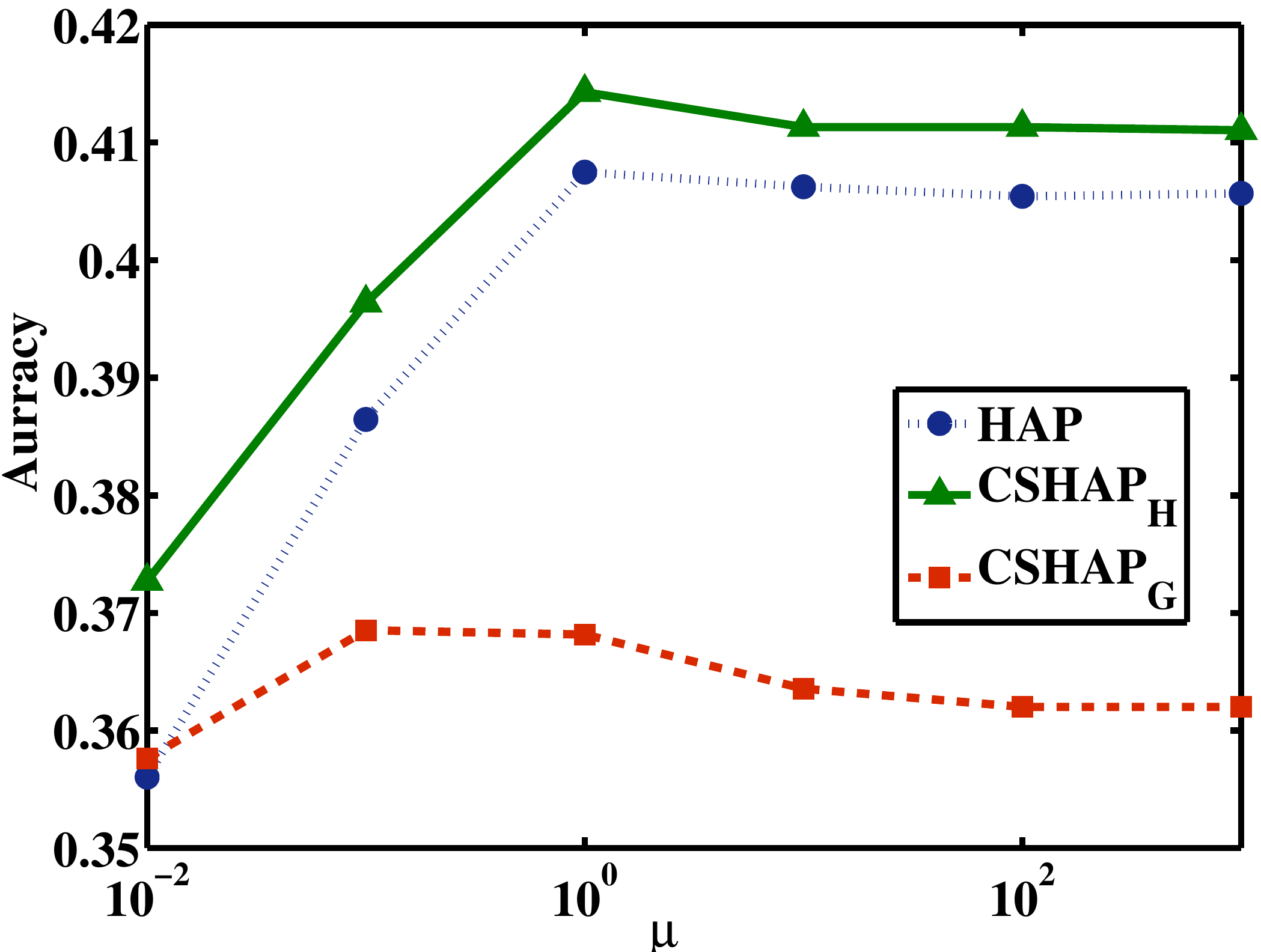}
}
\centering
\subfigure[CUB]{
\centering
\includegraphics[scale=0.25]{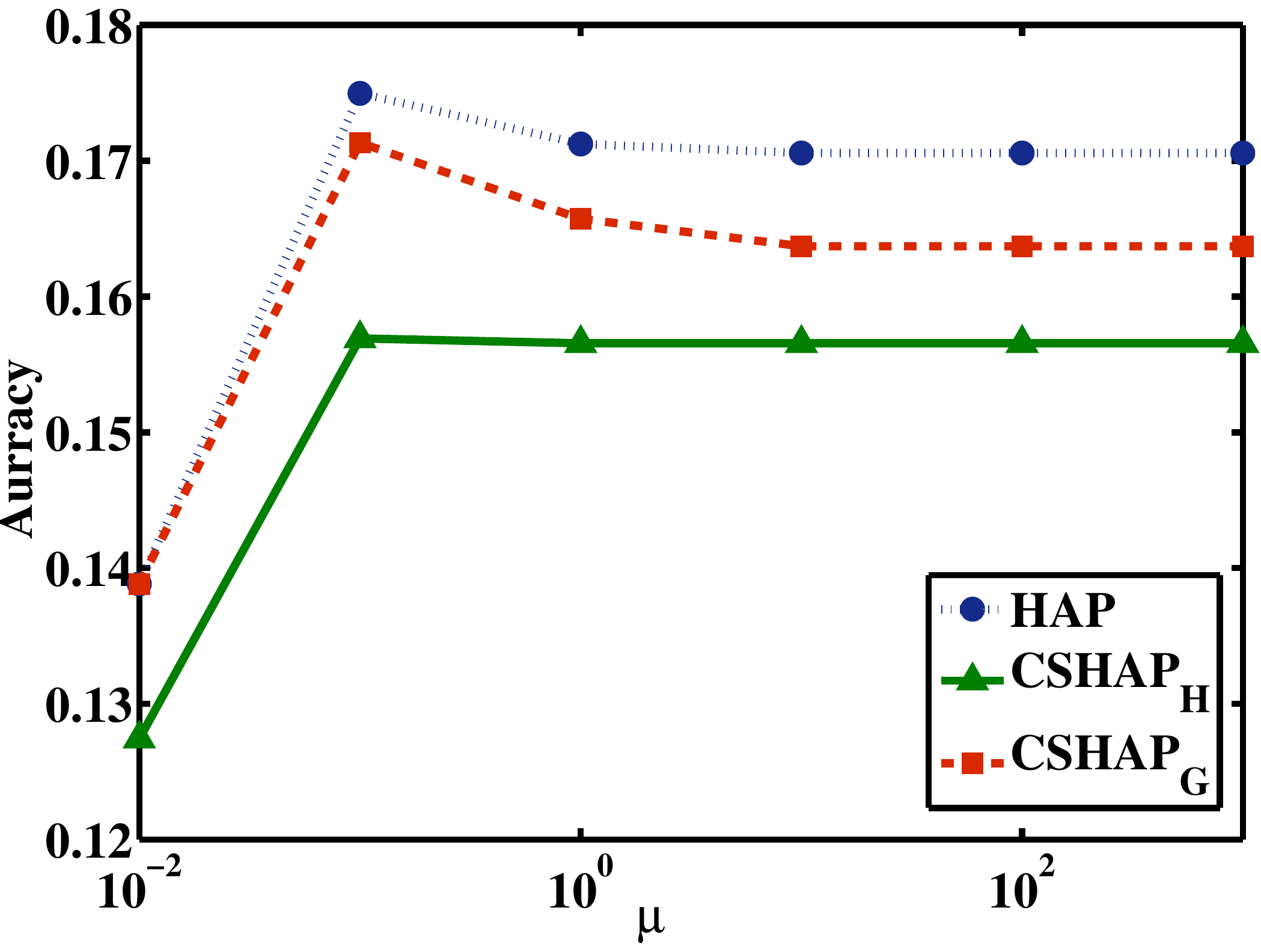}
}
\caption{The influences of $\mu$ to the ZSL accuracies.}
\label{mu}
\end{figure*}
\begin{figure*}[!h]
\setlength{\belowcaptionskip}{-0.4cm}
\centering
\subfigure[AWA]{
\centering
\includegraphics[scale=0.25]{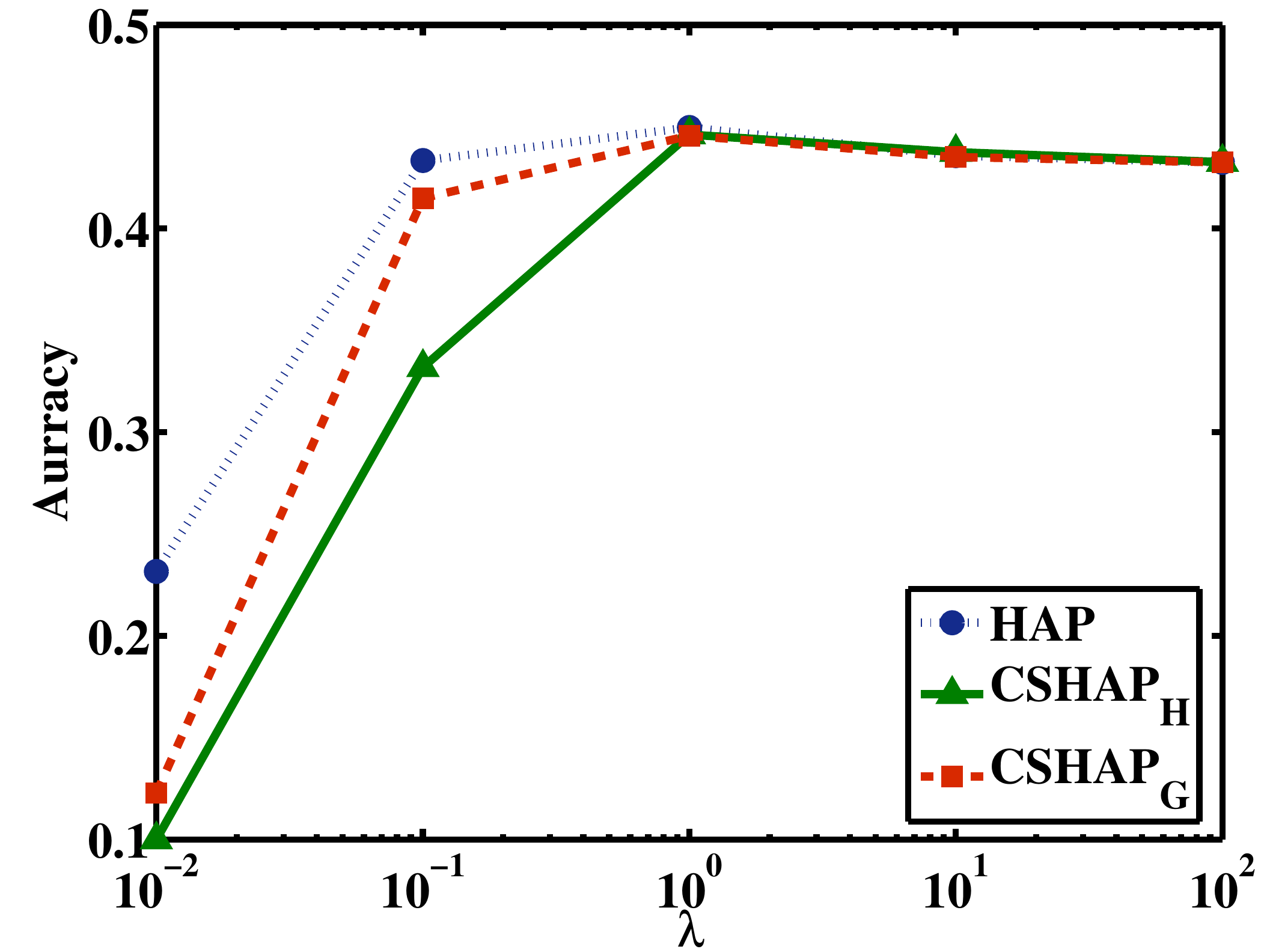}
}
\centering
\subfigure[USAA]{
\centering
\includegraphics[scale=0.25]{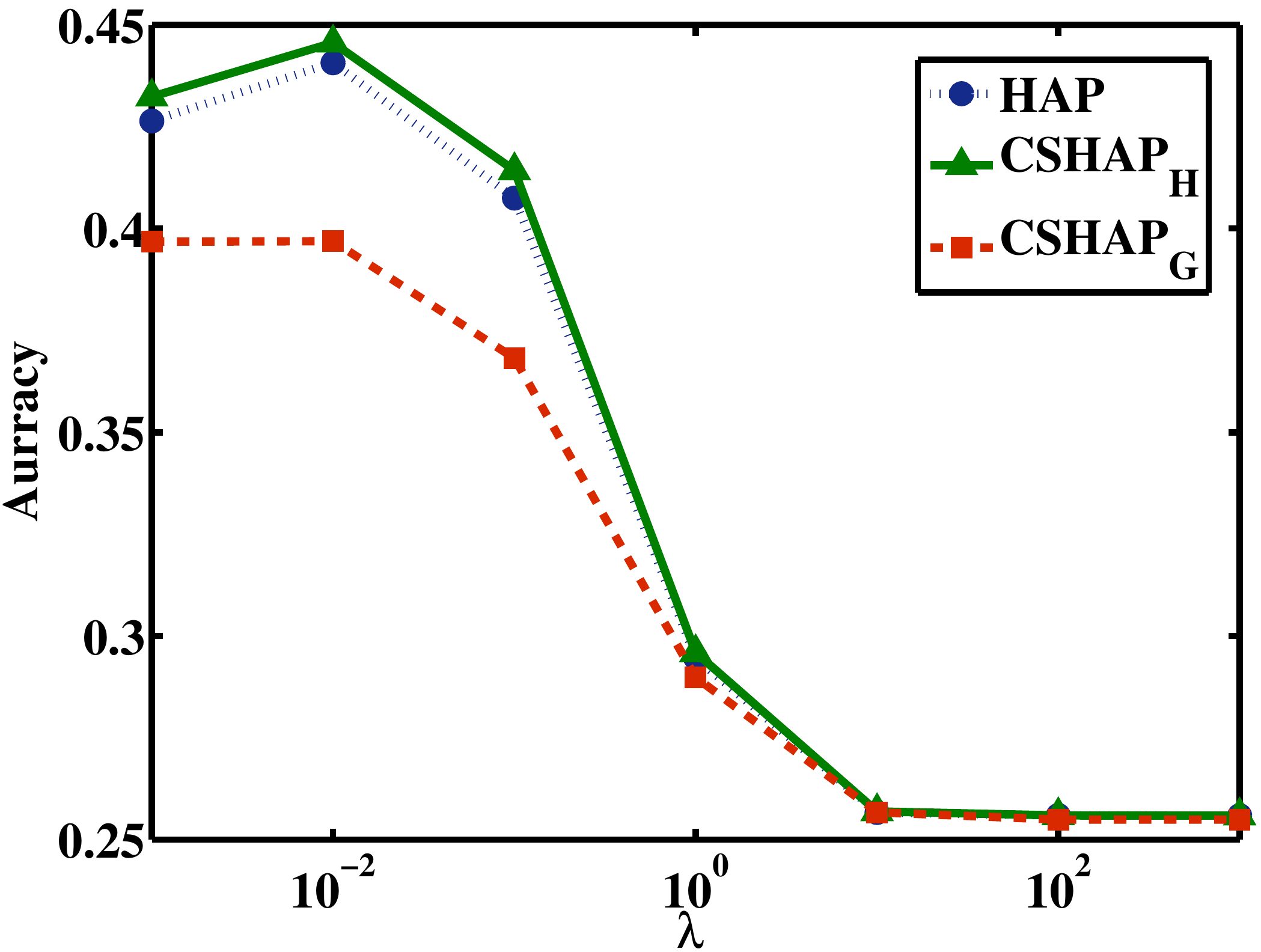}
}
\centering
\subfigure[CUB]{
\centering
\includegraphics[scale=0.25]{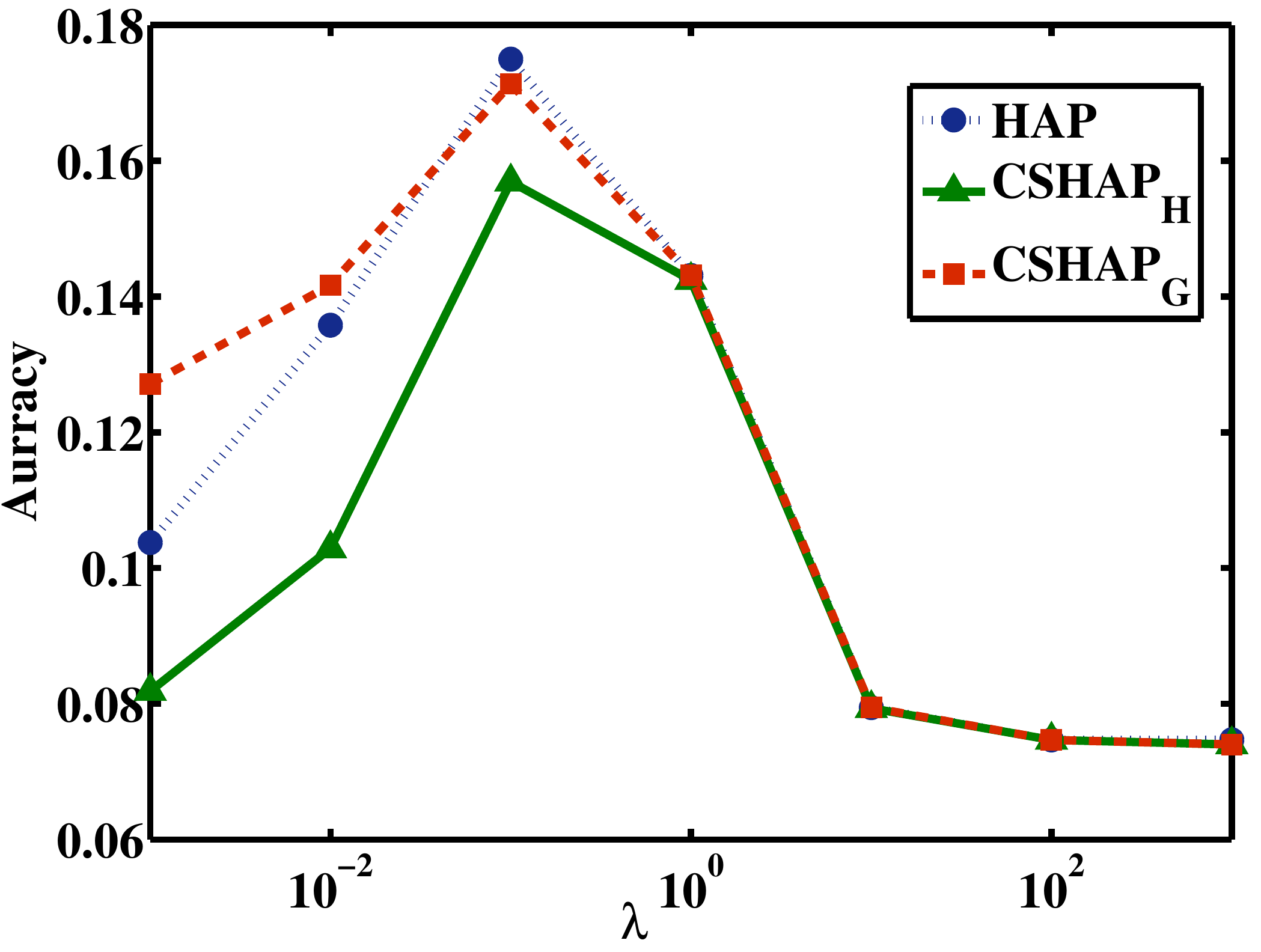}
}
\caption{The influences of $\lambda$ to the ZSL accuracies.}
\label{lambda}
\end{figure*}
\begin{figure*}[h]
\setlength{\belowcaptionskip}{-0.4cm}
\centering
\subfigure[AWA]{
\centering
\includegraphics[scale=0.25]{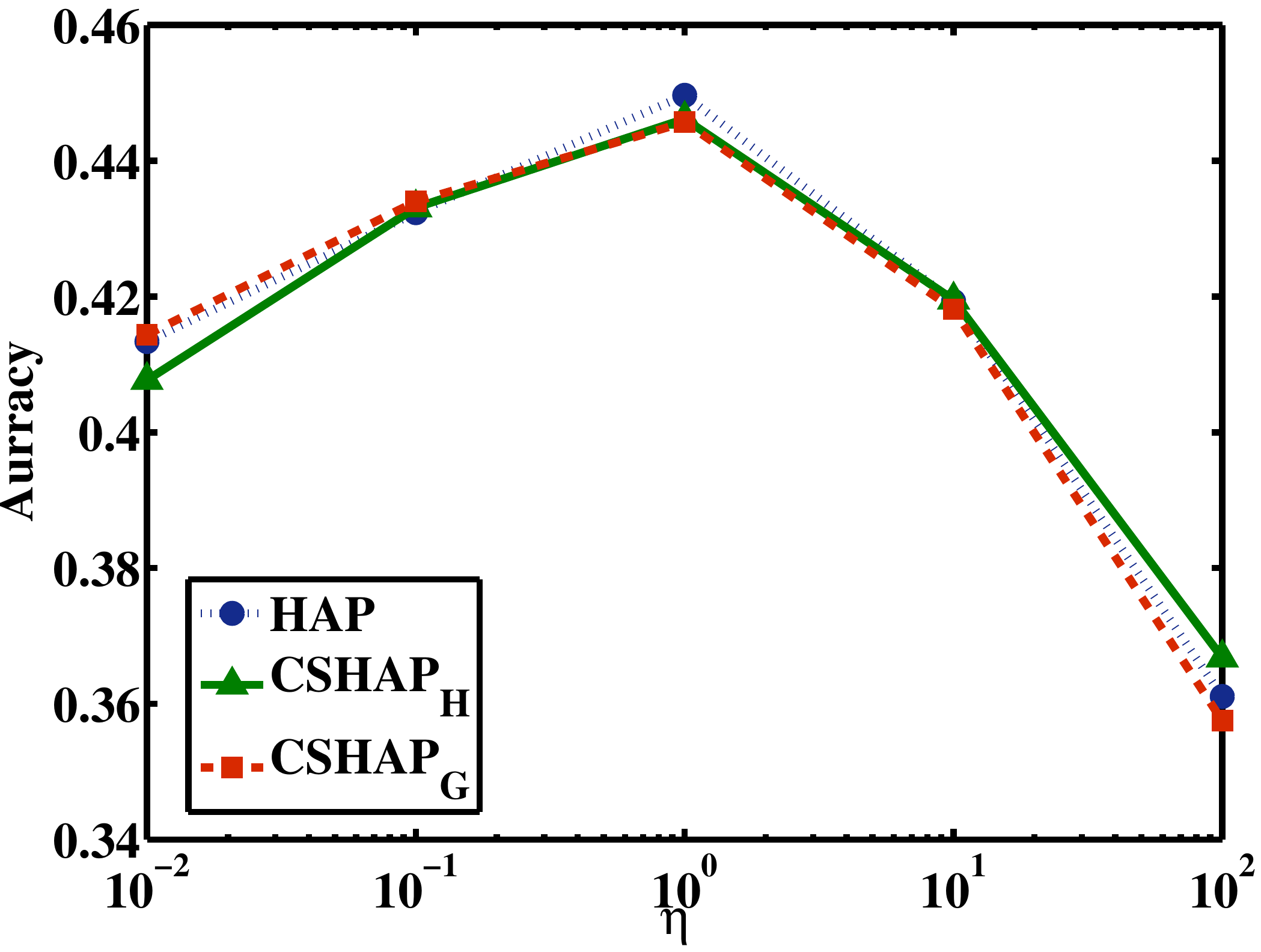}
}
\centering
\subfigure[USAA]{
\centering
\includegraphics[scale=0.25]{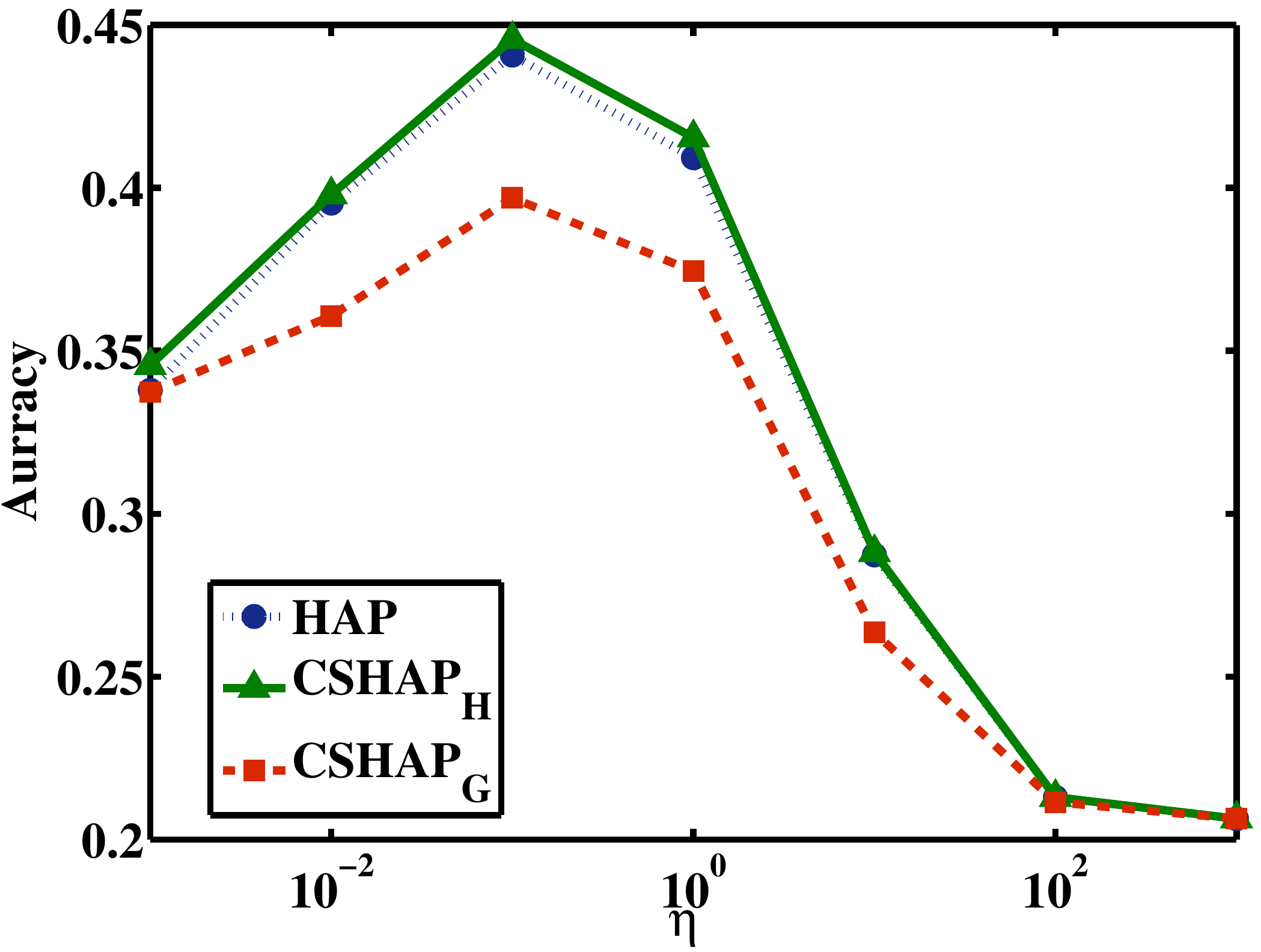}
}
\centering
\subfigure[CUB]{
\centering
\includegraphics[scale=0.25]{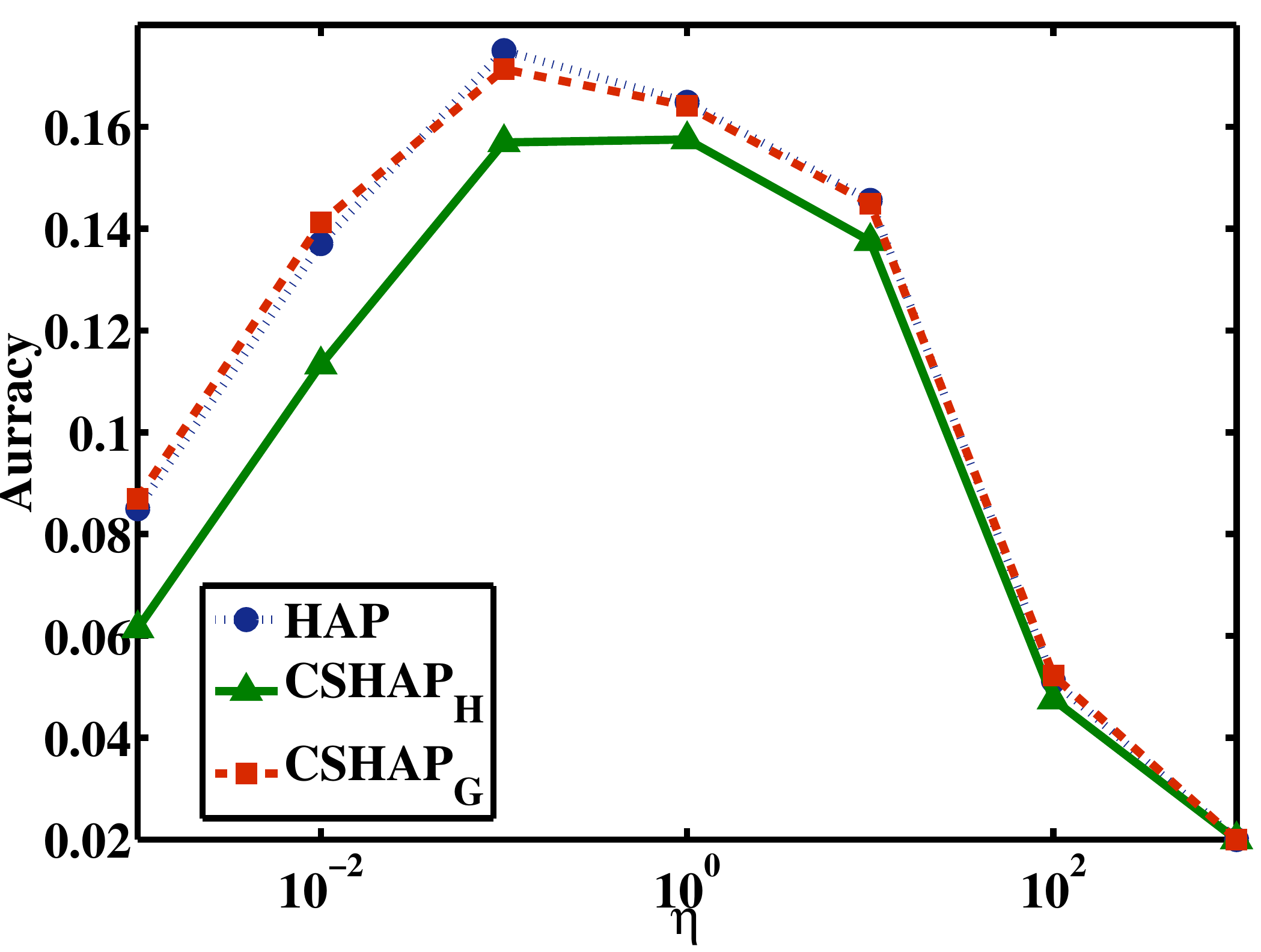}
}
\caption{The influences of $\eta$ to the ZSL accuracies.}
\label{eta}
\end{figure*}
\begin{figure*}[h]
\setlength{\belowcaptionskip}{-0.4cm}
\centering
\subfigure[AWA]{
\centering
\includegraphics[scale=0.25]{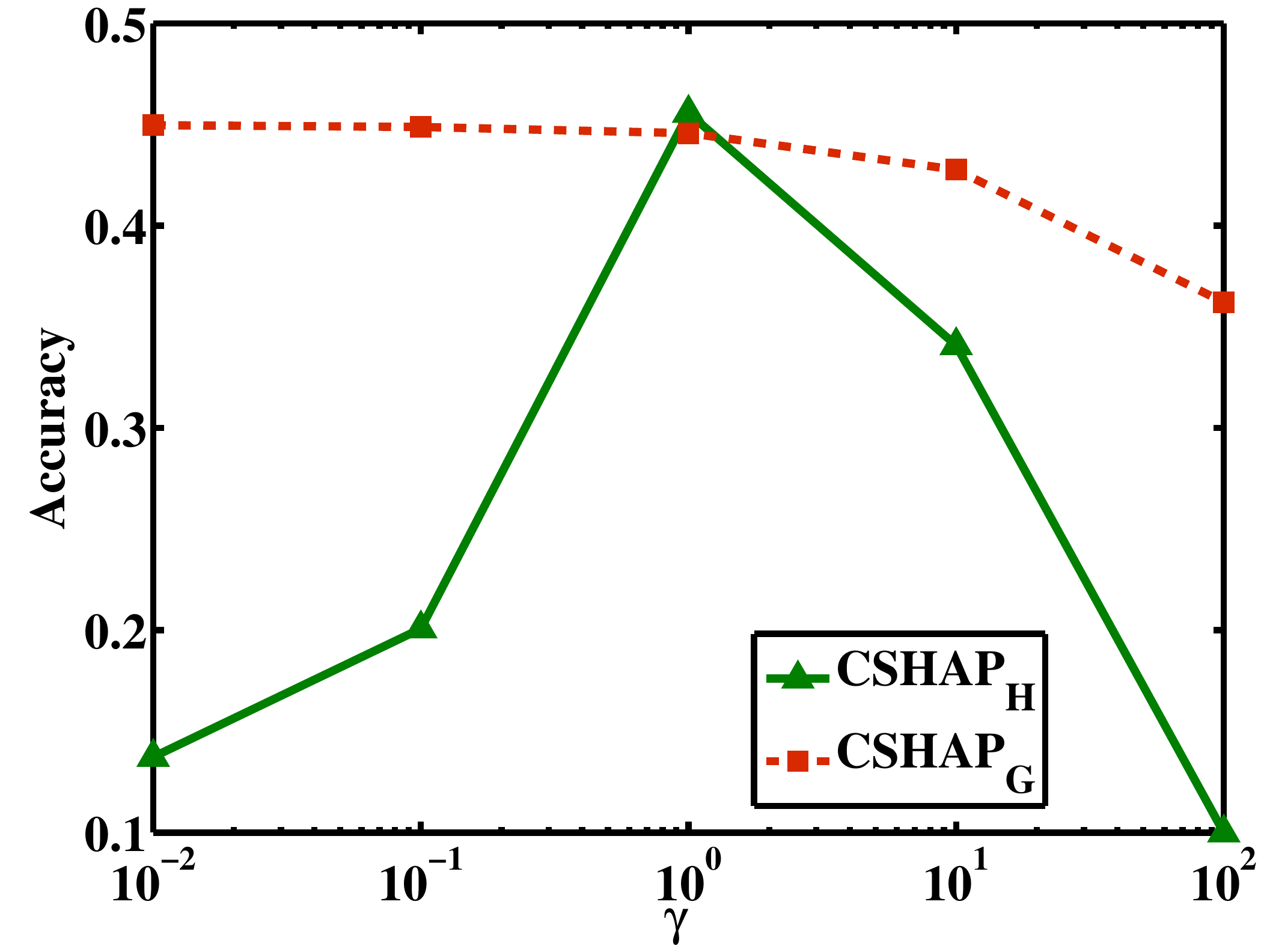}
}
\centering
\subfigure[USAA]{
\centering
\includegraphics[scale=0.25]{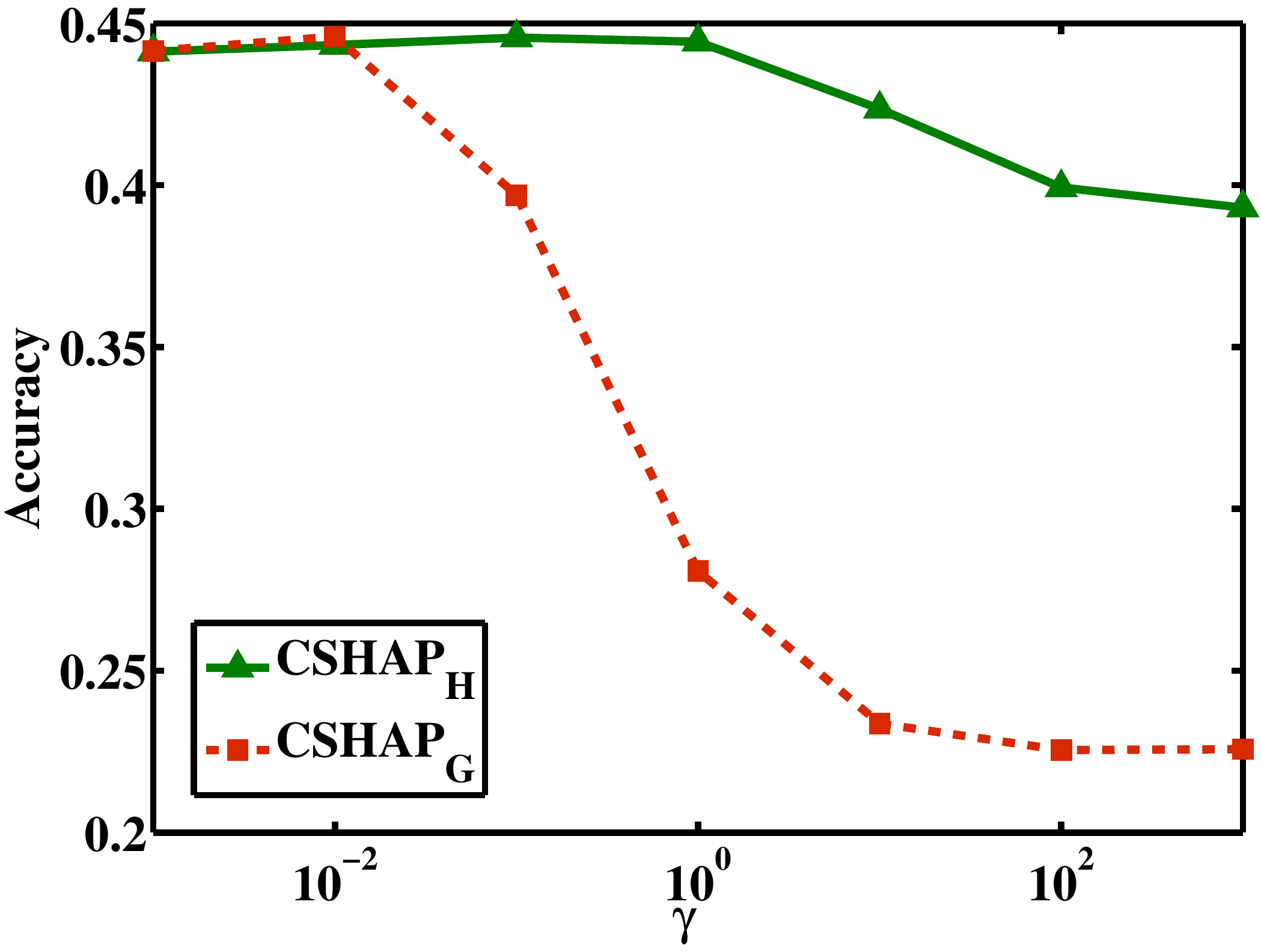}
}
\centering
\subfigure[CUB]{
\centering
\includegraphics[scale=0.25]{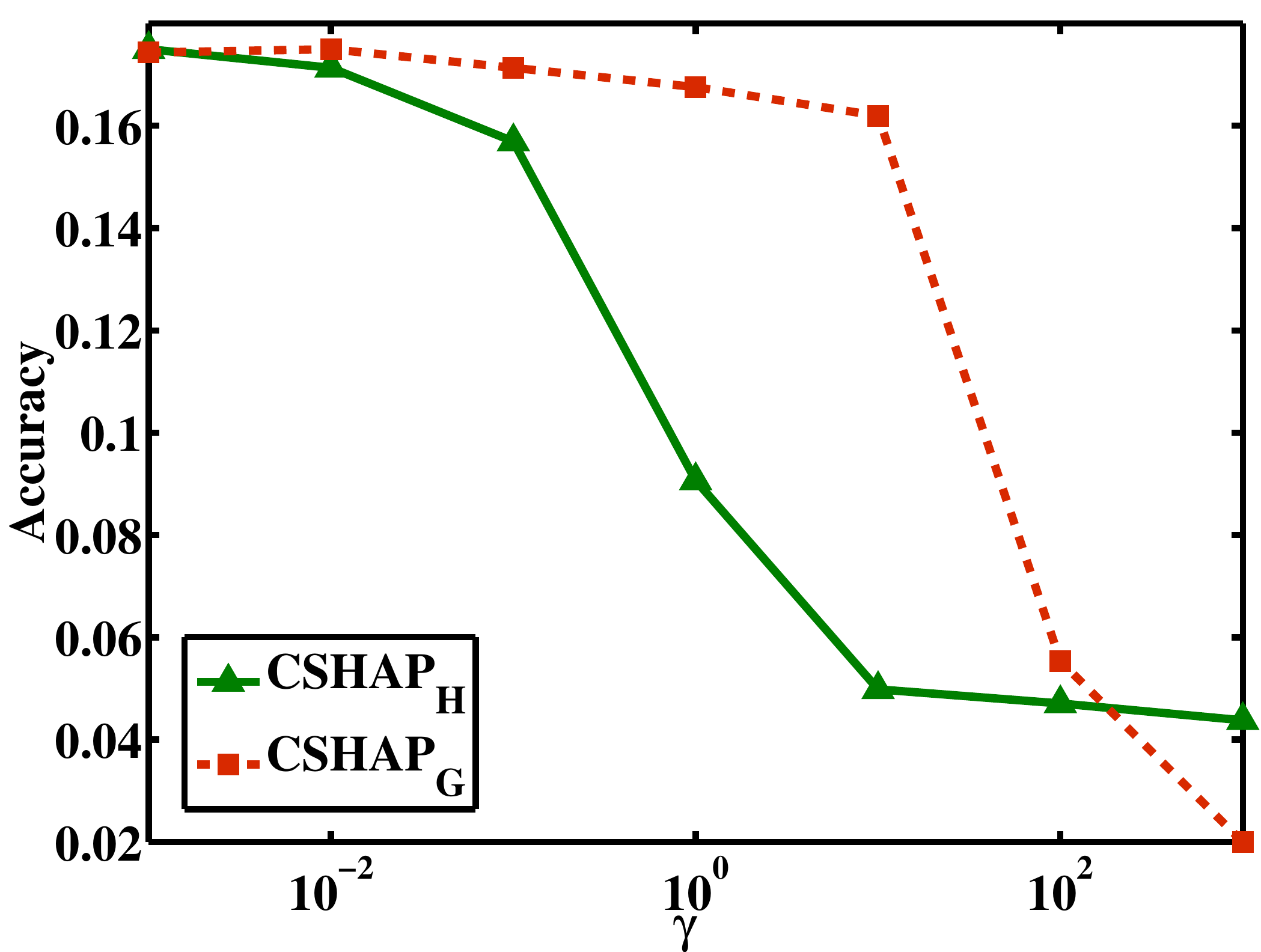}
}
\caption{The influences of $\gamma$ to the ZSL accuracies.}
\label{gamma}
\end{figure*}
\begin{figure*}[h]
\setlength{\belowcaptionskip}{-0.4cm}
\centering
\subfigure[AWA]{
\centering
\includegraphics[scale=0.25]{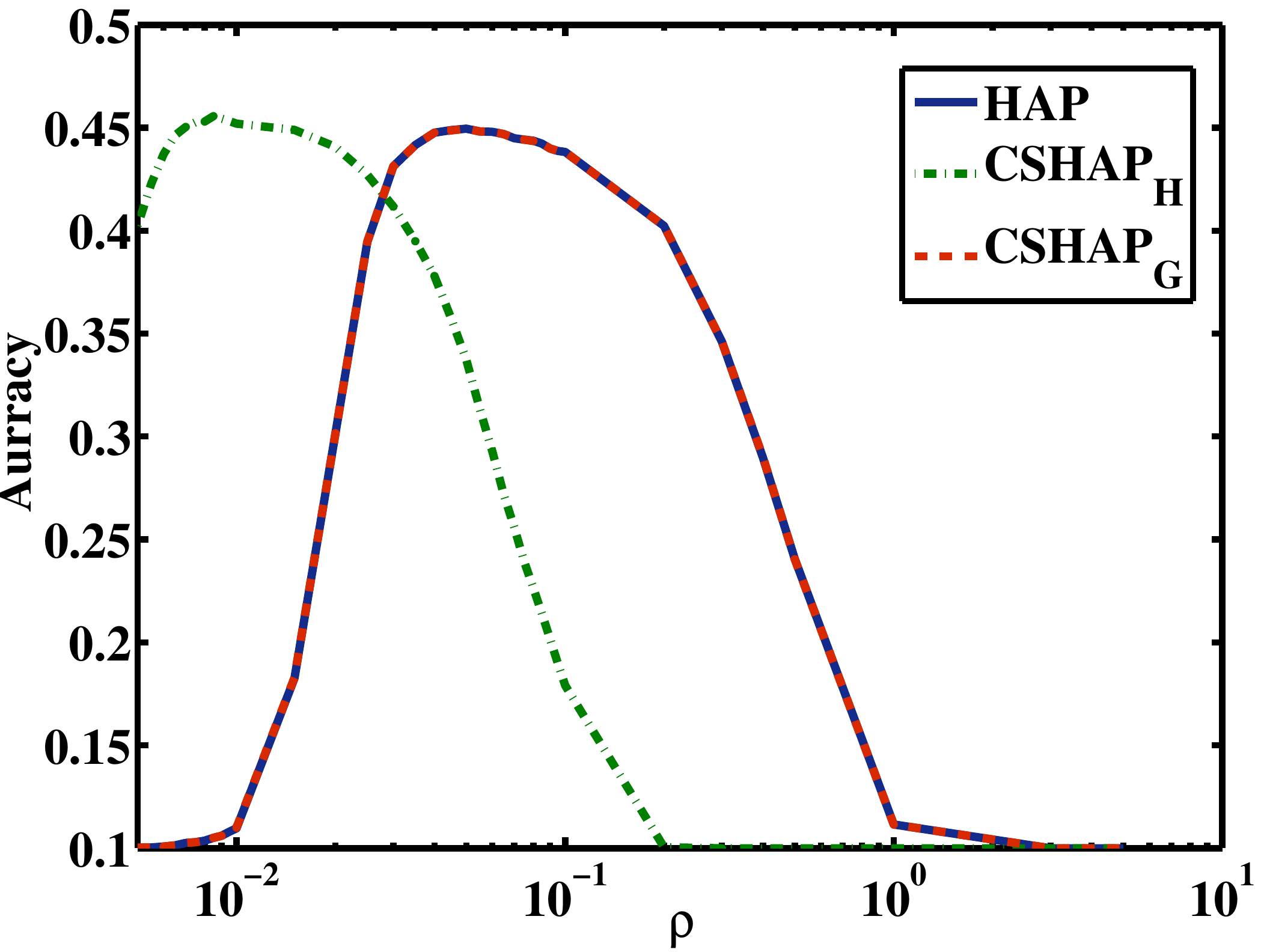}
}
\centering
\subfigure[USAA]{
\centering
\includegraphics[scale=0.25]{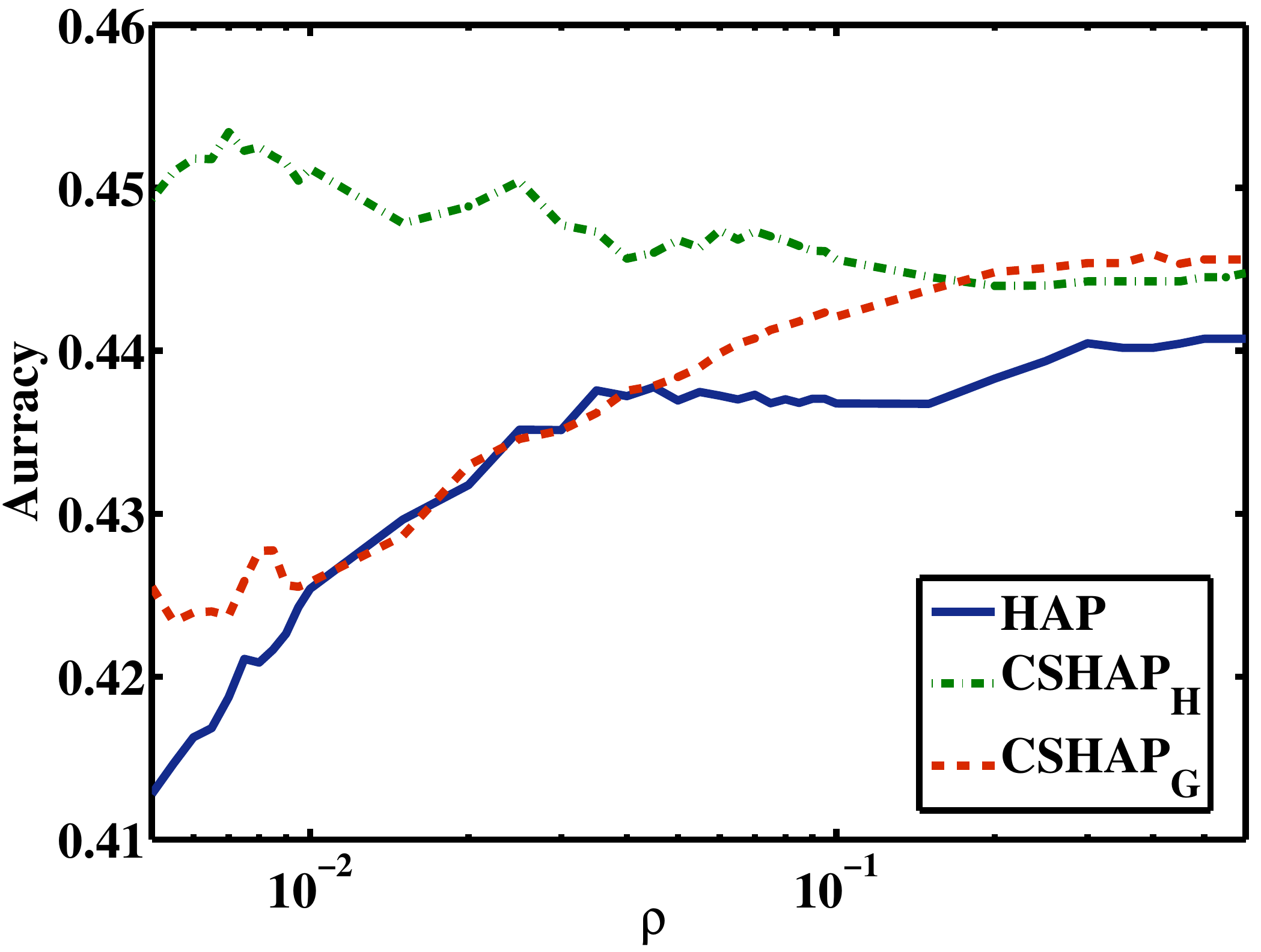}
}
\centering
\subfigure[CUB]{
\centering
\includegraphics[scale=0.25]{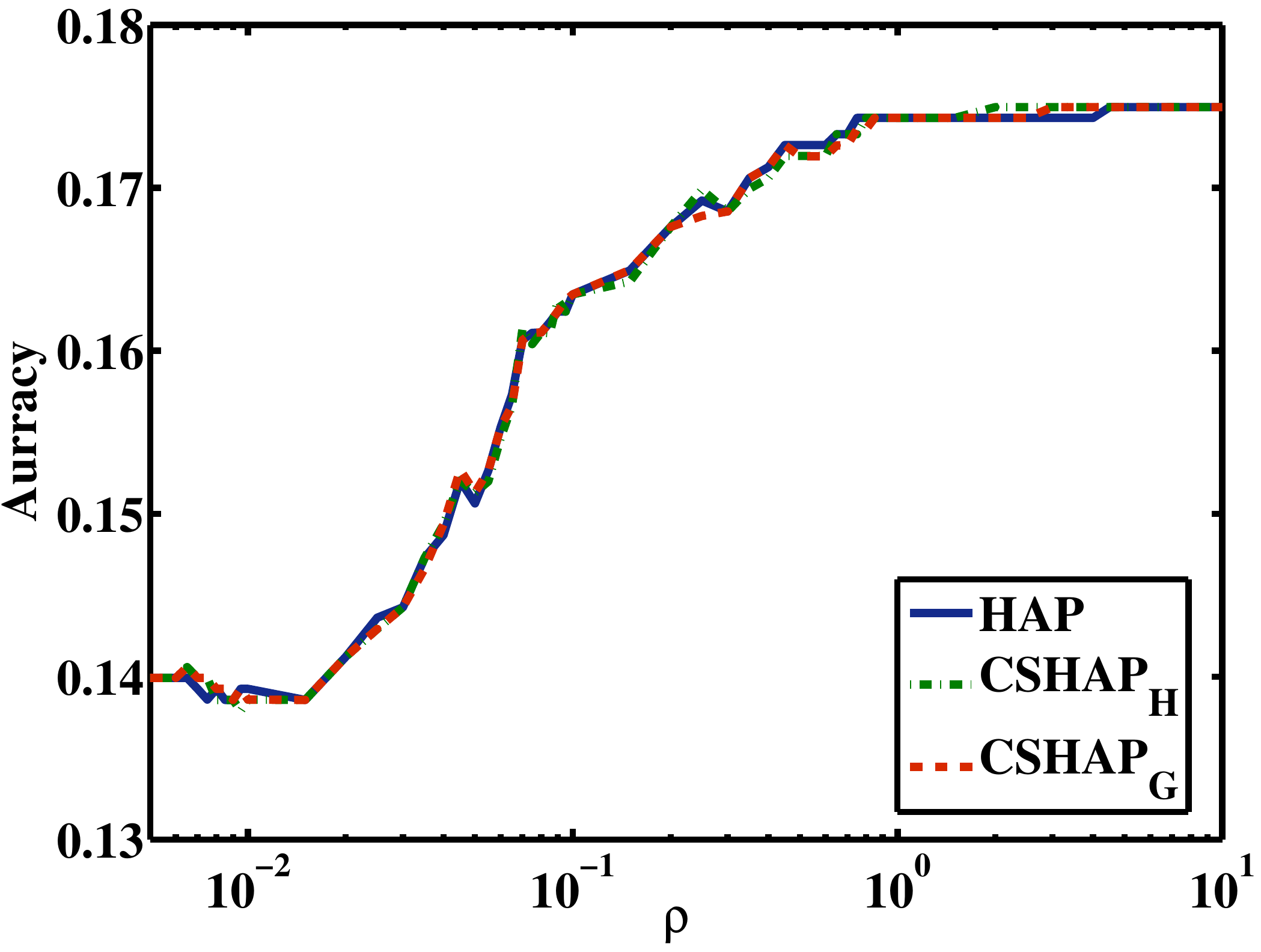}
}
\caption{The influences of $\rho$ to the ZSL accuracies.}
\label{rho}
\end{figure*}

}
\end{document}